\documentclass[10pt,twocolumn,letterpaper]{article}

\usepackage[T1]{fontenc}         
\usepackage[pagenumbers]{wacv}   

\usepackage{tikz}
\usetikzlibrary{shapes.geometric, arrows.meta, positioning, calc, fit, backgrounds, decorations.pathreplacing}

\newcommand{\method}{ARC-SORT}

\definecolor{wacvblue}{rgb}{0.21,0.49,0.74}
\usepackage[pagebackref,breaklinks,colorlinks,allcolors=wacvblue]{hyperref}

\def\wacvPaperID{}
\def\confName{WACV}
\def\confYear{2027}

\title{ARC-SORT: Adaptive, Robust, Compensated Multi-Object Tracking\\with Interacting Multiple Models}

\author{Girish Kumar Adari\\
Athlitix Inc.\\
{\tt\small girish@athlitix.com}}

\begin{document}
\maketitle
\begin{abstract}
Multi-object tracking in sports broadcasts is hard for two reasons:
athletes move fast and change direction abruptly, and the camera pans
and zooms at the same time. We present \method{} (\textbf{A}daptive,
\textbf{R}obust, \textbf{C}ompensated SORT), a tracking-by-detection
tracker that trains nothing and runs in real time. Its design rule:
the tracker measures its own uncertainty, and that measurement decides
how it acts. An Interacting Multiple Model (IMM) Kalman filter gives
every track a predictive covariance, and we turn it into a per-track
search region: the less reliable the motion prediction, the larger the
region in which a detection may match (uncertainty-adaptive Expansion
IoU, EIoU). On identical inputs, this continuous coupling is what
matters: the same covariance used as a hard Mahalanobis gate hurts at
every threshold we swept, and richer covariance summaries add nothing
measurable. When two tracks cross, the assignment cost itself reports
whether the decision was close. \method{} keeps such pairs under
review and applies at most one corrective identity swap once
post-crossing appearance evidence shows the first decision was wrong;
this adds $+1.0$ HOTA (higher order tracking accuracy) on both the
SportsMOT and DanceTrack test sets at under a millisecond per frame.
For players who leave the frame for seconds, motion extrapolation
carries no information, so our offline re-identification drops the
usual motion gate and links tracklets by appearance alone. \method{}
reaches $81.82$ HOTA on the SportsMOT test set with the fewest identity
switches of the 69 leaderboard entries as of August 2026 and tied-third
multi-object tracking accuracy (MOTA, $97.09$), and $62.1$ HOTA on
DanceTrack test with the same association settings. On Deep-EIoU's
public detections and embeddings it reaches $80.31$ HOTA, $+3.1$ over
Deep-EIoU. It runs at about 50 frames per second on a laptop RTX~3060
GPU. Code will be released.
\end{abstract}


\section{Introduction}
\label{sec:intro}

Tracking-by-detection splits multi-object tracking (MOT) into two
steps: a detector finds object boxes in every frame, and an
association step links those boxes across frames into tracks that
keep their identities. The SORT family~\cite{bewley2016sort}
performs the second step with a Kalman filter, which predicts where
each track will be, and a matching cost, usually the intersection
over union (IoU) between predicted and detected boxes, solved as a
linear assignment. This paper is about the association step only:
\method{} takes the output of a fixed detector and trains nothing.

Sports broadcasts concentrate the two failure modes of the SORT
paradigm into one domain: athletes sprint,
cut, and collide, violating the Kalman filter's constant-velocity
assumption, while the broadcast camera pans and zooms on top. The dominant response is to make association \emph{more tolerant},
enlarging boxes by a fixed margin~\cite{yang2023cbiou},
schedule~\cite{huang2024deepeiou}, or velocity
heuristic~\cite{wang2025memosort}, or else to \emph{learn} motion
per domain~\cite{lv2024diffmot, khanna2025sportmamba,
adzemovic2024movesort}. Meanwhile, two quantities that report how much the tracker should
trust itself are computed every frame and discarded: the filter's
predictive covariance says how far the motion prediction can be
trusted, yet overlap-cost pipelines either ignore it at association
time or consume it as a hard Mahalanobis veto~\cite{wojke2017deepsort}
(ground-plane costs~\cite{yi2024ucmctrack} abandon overlap entirely);
and the assignment's own cost margin says how nearly it preferred the
swapped pairing, yet nothing in the tracker's state distinguishes a
confident match from a coin flip.

This paper is built on a single principle: \textbf{a tracker should
let its own uncertainty govern its own decisions.} We instantiate it
twice, as search breadth and commitment firmness, and follow it to
its limit. At short horizons the motion covariance acts as
\emph{tolerance}, not veto: our \emph{uncertainty-adaptive
Expansion IoU} dilates each track's matching region by a continuous
function of its predictive standard deviation, supplied by an
Interacting Multiple Model (IMM)
filter~\cite{blom1988interacting, barshalom2001estimation}. At
crossings, the assignment margin governs \emph{identity commitment}:
a pair whose swap would cost the assignment almost nothing was
decided at the moment of worst evidence, so \method{} marks it
\emph{entangled}, snapshots pre-crossing appearance, accumulates a
permutation likelihood, and repairs the decision with at most one
corrective swap per episode~\cite{reid1979mht}. Past a multi-second
out-of-view gap (football's dominant failure; median camera exit
111~frames) the covariance has diverged and a search region wider
than the frame is no gate at all, so the offline re-identification
stage drops the motion gate and lets appearance alone link
tracklets.

Each part is established by measurement. A controlled
\emph{mechanism ladder} on identical cached inputs shows hard
Mahalanobis gating on the same covariance degrades association at
every threshold we swept, while richer uses add nothing measurable:
the scalar relative positional uncertainty is, empirically, a
sufficient statistic for association tolerance. Deferred commitment
is worth $+1.0$ HOTA on both test servers over the identical system
without it, and removing the merge's motion veto is worth $+4.9$ HOTA
on SportsMOT val and transfers to DanceTrack under a different
embedding model.

A tracker evaluated on a leaderboard is evaluated jointly with its
detector, so we anchor our comparison in a detector-deconfounded
protocol: on the \emph{exact public detections and embeddings} of
Deep-EIoU~\cite{huang2024deepeiou}, \method{} scores $80.31$
HOTA (higher order tracking accuracy~\cite{luiten2021hota}) versus
their published $77.20$, a $+3.1$ margin produced entirely
by association, the decisive share from removing the merge's motion
gate (Sec.~\ref{sec:exp_sota}). With our
own detector stack \method{} reaches $81.82$ on the SportsMOT test
set (rank~6 of 69 as of August 2026), with the fewest identity
switches of any entry (1{,}414) at tied-third MOTA ($97.09$) and,
despite the 30th-best detection accuracy (DetA), the 4th-best
identity F1 (IDF1): association, not detection, produces the result.

Our contributions are:
\begin{enumerate}
\item \textbf{Uncertainty-adaptive Expansion IoU:} to our knowledge
  the first IoU-family association tolerance set \emph{continuously}
  by the filter's own predictive covariance; training-free, one
  scalar per track per frame (Sec.~\ref{sec:adaptive_eiou}).
\item \textbf{A sufficient-statistic characterization:} a controlled
  ladder showing the scalar $\max(\sigma_x/w,\,\sigma_y/h)$ captures
  everything the covariance offers for association, and that
  veto-style uses of the same signal hurt
  (Secs.~\ref{sec:alternatives}, \ref{sec:exp_ladder}).
\item \textbf{Deferred identity commitment:} the assignment's
  permutation margin as an online ambiguity signal, converted by a
  two-hypothesis corrective swap into $+1.0$ test HOTA on two
  benchmarks at sub-millisecond cost (Secs.~\ref{sec:arbitration},
  \ref{sec:exp_arbitration}).
\item \textbf{The long-horizon corollary:} at multi-second
  out-of-view gaps the motion prior is the binding constraint;
  removing it from the offline merge is benchmark- and
  embedding-general, with every comparison anchored input-identical
  (Secs.~\ref{sec:merge}, \ref{sec:exp_merge}, \ref{sec:exp_sota}).
\end{enumerate}


\section{Related Work}
\label{sec:related}

\paragraph{Tracking-by-detection.}
SORT~\cite{bewley2016sort} established the predict--associate loop;
successive work strengthened every stage: appearance and post-hoc
refinement~\cite{wojke2017deepsort, du2023strongsort},
low-confidence recovery~\cite{zhang2022bytetrack},
observation-centric drift correction~\cite{cao2023ocsort}, camera
compensation~\cite{aharon2022botsort}, ground-plane
modeling~\cite{yi2024ucmctrack}, adaptive appearance
fusion~\cite{maggiolino2023deep_ocsort}, weak
cues~\cite{yang2024hybridsort}, and track-centric
assignment~\cite{shim2025tracktrack}. Across this family, the filter's
predictive covariance is either ignored at association time or, in
practice, consumed solely as a hard Mahalanobis
gate~\cite{wojke2017deepsort}; the association cost itself is
uncertainty-blind, and the assignment is \emph{commitment-blind}:
every match is final the frame it is made.

\paragraph{Expanded and buffered matching.}
Closest to \method{} are methods that enlarge boxes before computing
overlap: by a fixed proportional margin~\cite{yang2023cbiou}, a
fixed iterative schedule~\cite{huang2024deepeiou} (discarding the
Kalman filter), with a harmonic cost
mean~\cite{granhenriksen2024hmsort}, or switched by a binary
velocity threshold atop a trained long short-term memory (LSTM)
gated filter~\cite{wang2025memosort}. In every case the factor is fixed or
discretely switched; none derives it from the filter's own predictive
covariance. Ours is a continuous function of that covariance, wide
exactly when the prediction is unreliable, and requires no
training.

\paragraph{Uncertainty as a signal.}
In detection, probabilistic-IoU box
models~\cite{llerena2021probiou, wang2020jiou, kirsten2023cell}
score overlap between distributions rather than boxes;
UTrack~\cite{solanocarrillo2024utrack} propagates \emph{detector}
uncertainty into association, complementary to the
\emph{motion-model} uncertainty we exploit. IMM
filtering~\cite{blom1988interacting, mazor1998interacting,
barshalom2001estimation} has long supplied calibrated predictive
covariances in radar tracking, where the probabilistic and joint
probabilistic data association filters (PDAF/JPDAF)~\cite{barshalom1975pdaf,
fortmann1983jpdaf} spend them continuously inside a soft filter
update. Visual MOT's hard-assignment, overlap-cost pipelines use the
covariance only as a Mahalanobis veto~\cite{wojke2017deepsort};
covariance as a continuous overlap \emph{tolerance} under one-to-one
assignment is, to our knowledge, new.

\paragraph{Deferred data association.}
Multiple hypothesis tracking (MHT)~\cite{reid1979mht} answers ambiguity by not deciding; PDAF/JPDAF
(above) soften the assignment itself. Vision revisited both, in
pruned MHT with appearance~\cite{kim2015mht}, $m$-best
JPDA~\cite{rezatofighi2015jpda}, and near-online
revision~\cite{choi2015nomt}, but the SORT lineage dropped
deferral: it commits every identity at the crossing instant,
precisely when occlusion makes evidence worst, and revision survives
only as offline refinement~\cite{sun2024gta, du2023strongsort}. Our
mechanism (Sec.~\ref{sec:arbitration}) is a bounded revival inside
the causal stream: the assignment's own permutation margin selects
the rare entangled pairs, two hypotheses each, at most one corrective
swap per episode; to our knowledge no SORT-family tracker on these
benchmarks revises a committed identity within the causal stream at
SORT cost.

\paragraph{Learned motion and association.}
The alternative is to learn: motion
networks~\cite{qin2023motiontrack, lv2024diffmot,
huang2024mambamot, hu2024trackssm, khanna2025sportmamba}, deep
Kalman replacements~\cite{adzemovic2024movesort,
adzemovic2024deepmovesort}, learned
association~\cite{somers2025cameltrack, adzemovic2025tdlp},
graph-based online/offline tracking~\cite{missaoui2025nooougat},
and end-to-end transformers~\cite{zeng2022motr, gao2025motip},
all trained per domain, inheriting their training distribution;
\method{} adapts online through the filter, training-free and
detector-independent.

\paragraph{Offline tracklet association.}
Closest to our long-horizon stage, GTA~\cite{sun2024gta} refines
Deep-EIoU offline, globally re-linking tracklets under appearance
\emph{plus spatio-temporal} constraints (81.0 on SportsMOT). Our
merge differs in what it removes: it drops the motion constraint at
long gaps, the binding factor per our ablation
(Sec.~\ref{sec:exp_merge}), and links on appearance alone.

\paragraph{Foundation-model tracking and benchmarks.}
At a different compute tier, selective mask
propagation~\cite{holmberg2026selective} dispatches a SAM~3 video
segmenter only where an assignment-uncertainty signal fires (87.2 on
SportsMOT, with global association): our premise, uncertainty decides
where effort is spent, at a fraction of the cost.
SportsMOT~\cite{cui2023sportsmot} combines fast, non-linear athlete
motion with panning broadcast cameras; DanceTrack~\cite{sun2022dancetrack}
isolates non-linear motion under a largely static camera, so the pair
is a controlled test of motion and camera components.


\section{Method}
\label{sec:method}

\begin{figure*}[t]
\centering
\resizebox{0.94\textwidth}{!}{
\begin{tikzpicture}[
    node distance=0.4cm and 0.4cm,
    box/.style={rectangle, draw, rounded corners=2pt, minimum height=0.85cm, minimum width=1.45cm, align=center, font=\small},
    mainbox/.style={box, fill=blue!10, draw=blue!55, line width=0.6pt},
    novelbox/.style={box, fill=orange!18, draw=orange!75, line width=0.9pt},
    offbox/.style={box, fill=green!10, draw=green!55, line width=0.6pt},
    iobox/.style={box, fill=gray!12, draw=gray!55, line width=0.6pt},
    arrow/.style={-{Stealth[length=5pt]}, thick},
    darrow/.style={-{Stealth[length=5pt]}, thick, dashed},
    lbl/.style={font=\scriptsize\itshape, text=gray!65!black},
]
\node[iobox] (dets) {Detections $\mathcal{D}_t$};
\node[mainbox, right=of dets] (split) {BYTE split\\$\mathcal{D}_H,\mathcal{D}_L$};
\node[mainbox, right=of split] (imm) {IMM predict\\CV$\,\|\,$CA$\,\|\,$CT};
\node[mainbox, right=of imm] (cmc) {CMC warp\\(similarity)};
\node[novelbox, right=of cmc] (unc) {$S=\hat P_{xy}{+}R$\\$e=\mathrm{clip}(\alpha\,\tfrac{\sigma}{\text{size}})$};
\node[mainbox, right=of unc] (a1) {Stage 1\\EIoU$_e$+OCM+app};
\node[mainbox, right=of a1] (a2) {BYTE 2\\IoU};
\node[mainbox, right=of a2] (a3) {OCR\\IoU (+app)};
\node[mainbox, right=of a3] (upd) {Update\\ORU$^{\text{CMC}}$};
\node[novelbox, right=of upd] (arb) {Defer/swap\\margin$\,\to\,\ell$};
\node[iobox, below=0.5cm of imm] (frame) {Frame $I_t$};
\node[mainbox, below=0.5cm of a1] (reid) {OSNet embeddings (cached)};
\draw[arrow] (dets) -- (split);
\draw[arrow] (split) -- (imm);
\draw[arrow] (imm) -- (cmc);
\draw[arrow] (cmc) -- (unc);
\draw[arrow] (unc) -- (a1);
\draw[arrow] (a1) -- (a2);
\draw[arrow] (a2) -- (a3);
\draw[arrow] (a3) -- (upd);
\draw[arrow] (upd) -- (arb);
\draw[darrow] (frame.north) -| (cmc.south);
\draw[darrow] (frame.east) -- (reid.west);
\draw[arrow] (reid.north) -- (a1.south);
\draw[darrow] (split.south) |- ([yshift=-1.6cm]split.south) -| (a2.south);
\node[lbl] at ([yshift=-1.78cm]$(split.south)!0.35!(a2.south)$) {$\mathcal{D}_L$ (low confidence)};
\node[lbl, above=0.06cm of unc] {\textbf{short horizon: covariance as tolerance}};
\node[offbox, below=1.9cm of a1] (interp) {Interpolate\\gaps $\le 10$};
\node[iobox, left=of interp] (tracklets) {Tracklets};
\node[offbox, right=of interp] (proto) {Prototype\\$\bar f$ (mean, $\ell_2$)};
\node[novelbox, right=of proto] (merge) {Merge: $\cos\ge0.85$, gap$\,\le300$\\\textbf{motion gate removed}};
\node[iobox, right=of merge] (out) {Final tracks};
\draw[arrow] (upd.south) -- ++(0,-1.85) -| (tracklets.north);
\draw[arrow] (tracklets) -- (interp);
\draw[arrow] (interp) -- (proto);
\draw[arrow] (proto) -- (merge);
\draw[arrow] (merge) -- (out);
\node[lbl, below=0.06cm of merge] {\textbf{long horizon: appearance only; extrapolation is information-free}};
\node[lbl, left=0.1cm of tracklets] {\emph{offline, per sequence:}};
\end{tikzpicture}
}
\caption{\textbf{\method{}: one principle, twice instantiated, once taken to its limit.} Online
(top): the IMM predictive covariance, warped by camera-motion
compensation, sets a per-track expansion $e_i$ dilating the matching
region exactly when the prediction is unreliable; near-ambiguous
match pairs (low permutation margin) enter a deferred two-hypothesis
state, settled by at most one corrective identity swap per episode
(labels only, filter states untouched; orange). Offline (bottom): across multi-second
gaps tracklets link by appearance prototypes alone, motion veto
removed (orange). OCM, OCR, ORU: observation-centric momentum,
recovery, and re-update (Secs.~\ref{sec:cmc}, \ref{sec:association}).}
\label{fig:architecture}
\end{figure*}
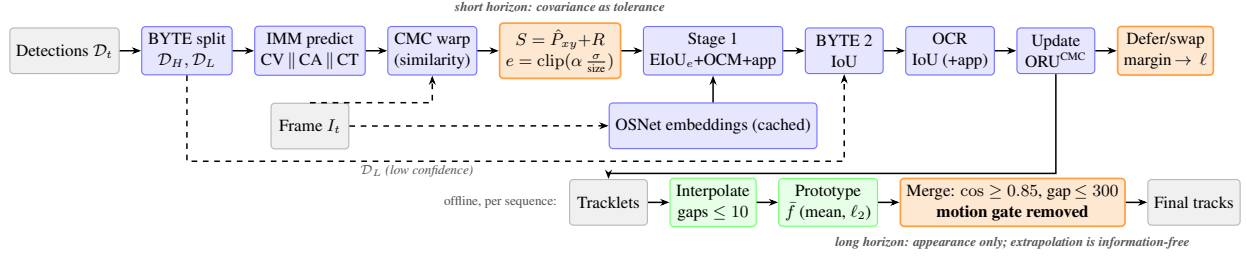

\method{} applies one rule three times (Fig.~\ref{fig:architecture}).
Online, frame by frame: each track's matching region is dilated in
proportion to its own predictive covariance
(Sec.~\ref{sec:adaptive_eiou}), and near-ambiguous match pairs,
identified by the assignment's own margin, are held in a deferred
two-hypothesis state and corrected by a single identity swap when
later evidence rules the first decision wrong
(Sec.~\ref{sec:arbitration}). Offline, per sequence: tracklet
fragments separated by multi-second gaps, where motion information
has run out, are linked by appearance alone, with the customary
motion veto removed (Sec.~\ref{sec:merge}).

\subsection{Online tracker overview}
\label{sec:overview}

At each frame, detections split by confidence at
$\tau_H{=}0.6$/$\tau_L{=}0.1$~\cite{zhang2022bytetrack}; each
track's IMM filter predicts (Sec.~\ref{sec:imm}); a camera-motion
estimate warps all predicted states into the current frame
(Sec.~\ref{sec:cmc}); the warped covariance sets each track's
expansion (Sec.~\ref{sec:adaptive_eiou}); three association stages
match detections to tracks (Sec.~\ref{sec:association}). Matched
tracks receive a filter update (replayed through stored camera
transforms after loss, Sec.~\ref{sec:cmc}); unmatched
high-confidence detections spawn tracks ($\ge 0.7$); tracks lost
over 30 frames are removed. After lifecycle management, an
arbitration pass reads the frame's assignment margins and manages
entanglement episodes (Sec.~\ref{sec:arbitration}); it can
relabel two tracks but never touches a filter state.

\subsection{IMM motion model}
\label{sec:imm}

\paragraph{Models.} All models operate on measurements
$\mathbf{z}=[c_x,c_y,w,h]^\top$. A constant-velocity (CV) model tracks
all four box coordinates and their velocities
($\mathbb{R}^8$); a constant-acceleration (CA) model adds
acceleration for the center only ($\mathbb{R}^{10}$); a
coordinated-turn (CT) model adds a turn rate $\omega$
($\mathbb{R}^9$) with the standard non-linear center
transition~\cite{barshalom2001estimation}, handled by an extended
Kalman filter (EKF) with an
analytic Jacobian; $|\omega|<10^{-6}$ reduces to CV.

\paragraph{Cycle.} The IMM cycle~\cite{blom1988interacting} runs
mixing, per-model prediction, log-space probability update, and
combination ($\Pi_{ii}=0.98$, initial $[0.7,0.15,0.15]$; a
${\sim}50$-frame adaptation constant). Mixing operates on the common
8-dimensional subspace; model-specific states pass through unmixed
with cross-covariances reset, a reduced-state approximation of
textbook IMM. The combined covariance
$\hat P=\sum_j \mu_j\big(G_jP_jG_j^\top +
(G_j\mathbf{x}_j-\hat{\mathbf{x}})(\cdot)^\top\big)$ includes the
between-model spread: disagreement among motion hypotheses
\emph{is} uncertainty, and it widens the expansion below.

\paragraph{Noise.} Process and measurement noise are diagonal and
size-dependent, following BoT-SORT~\cite{aharon2022botsort}:
position stds $w/20,h/20$, velocity stds $w/160,h/160$, recomputed
each step ($Q$ deliberately diagonal, not the discretized white-noise
form).

\subsection{Camera motion compensation (CMC) and re-update}
\label{sec:cmc}

A four-degree-of-freedom (4-DOF) similarity transform is estimated per frame pair by sparse
optical flow (1{,}000 corners, pyramidal Lucas--Kanade,
$2{\times}$ downscale, RANSAC, identity fallback). With
rotation-scale block $R$, translation $\mathbf{t}$, and
$s=\sqrt{|\det R|}$, every sub-model's state is warped (position
by $R(\cdot)+\mathbf{t}$, velocities by $R$, sizes by $s$), and
covariances by $P\!\leftarrow\!TPT^\top$.

\paragraph{CMC-corrected re-update.} OC-SORT's observation-centric
re-update (ORU)~\cite{cao2023ocsort} repairs filter drift after a
re-found track by replaying interpolated virtual observations across
the gap. But under a moving camera, pixel-space interpolation
traces the camera's path, not the object's. We store per-frame
transforms $A_t$; on re-match after gap $g$, the new measurement is
mapped to the frozen frame, interpolated there, and each virtual
observation mapped forward before a predict--warp--update replay:
\begin{equation}
\tilde{\mathbf{z}}_s = \Big(\textstyle\prod_{i\le s} A_i\Big) \circ
\big[(1{-}\tfrac{s}{g})\,\mathbf{z}_{\text{last}} +
\tfrac{s}{g}\,\mathbf{z}_{\text{new}}^{\text{frozen}}\big],
\quad s=1..g.
\label{eq:cmcoru}
\end{equation}
The corrected replay activates only when some gap-frame translation
reaches 1\,px; static-camera gaps use standard ORU.

\subsection{Uncertainty-adaptive Expansion IoU}
\label{sec:adaptive_eiou}

This is the paper's core mechanism. After prediction and CMC, each
track exposes its innovation covariance in measurement space,
$S=\hat P_{[1:4,1:4]}+R$, where $\hat P$ is the IMM combined
covariance (including between-model spread) and $R$ the
size-dependent measurement noise. From its positional diagonal we
form the \emph{relative positional uncertainty} and map it to an
expansion factor:
\begin{equation}
e_i \;=\; \operatorname{clip}\!\Big(\,\alpha \cdot
\max\!\Big(\frac{\sigma_x}{w},\ \frac{\sigma_y}{h}\Big),\;
e_{\min},\; e_{\max}\Big),
\label{eq:adaptive}
\end{equation}
with $\sigma_x=\sqrt{S_{11}}$, $\sigma_y=\sqrt{S_{22}}$, predicted
box size $(w,h)$, and $\alpha{=}2$, $e_{\min}{=}0.2$,
$e_{\max}{=}1.2$ throughout: the matching region covers two
predicted-center standard deviations in units of box size. Both boxes of a
pair grow symmetrically before computing IoU, each side by $e_i$
times that box's own extent~\cite{huang2024deepeiou}; $e_i{=}0$
reduces exactly to IoU. Unlike Deep-EIoU's global schedule, $e_i$ is
\emph{per track, per frame}: a track mid-maneuver casts a wide net;
a confidently tracked one keeps a tight gate. The mechanism is
training-free and costs one covariance read per track.

Fig.~\ref{fig:supp_mechanism} in the appendix traces $e_i$ live on a val sequence: it hugs
$e_{\min}$ under confident tracking, widens at maneuver onsets and
across missed detections, and snaps back on re-capture.


\subsection{Association pipeline}
\label{sec:association}

\paragraph{Stage 1.} High-confidence detections are assigned to
predicted tracks by maximizing
\begin{equation}
\text{score}_{nm} = \mathrm{EIoU}_{e_m}(d_n, p_m)
 + s^{\text{ocm}}_{nm} + 1.25\cdot \hat E_{nm},
\label{eq:cost}
\end{equation}
solved as a linear assignment. The observation-centric momentum (OCM)
term $s^{\text{ocm}}$~\cite{cao2023ocsort} is a signed alignment bonus
($\pm 0.1$, $\Delta t{=}3$ observation history); $\hat E$ is the
cosine between the detection embedding and the track's
confidence-trust exponential-moving-average (EMA) embedding
($\alpha_{\text{ema}}{=}0.95{+}0.05(1{-}\text{trust})$; 512-d
sports-domain OSNet~\cite{huang2024deepeiou}), masked to zero
wherever the spatial term is zero (appearance refines spatially
plausible matches but never creates one), at fixed weight $1.25$.
Matches require the spatial term to reach $0.25$.

\paragraph{Stage 2 (BYTE second pass).} Still-unmatched recently tracked targets
are matched to low-confidence detections by plain IoU
($\ge 0.5$)~\cite{zhang2022bytetrack}: no expansion, momentum, or
appearance.

\paragraph{Stage 3 (observation-centric recovery, OCR).} Remaining high-confidence detections are
compared against unmatched tracks' \emph{last observed} boxes
\cite{cao2023ocsort}; pairs must still reach IoU $\ge 0.25$, the
appearance term (unweighted here) only re-ranking spatially valid
recoveries.

\subsection{Deferred identity commitment}
\label{sec:arbitration}

Crossings force an identity decision at the moment of worst
evidence, and the assignment matrix already knows it. For matches
$a=(d_i,t_j)$ and $b=(d_k,t_l)$ accepted by the Stage-1 assignment of
Eq.~\eqref{eq:cost}, we read off the \emph{permutation margin} from
that stage's score matrix,
\begin{equation}
\Delta_{ab} \;=\;
\big(\text{score}_{ij}+\text{score}_{kl}\big)
-\big(\text{score}_{il}+\text{score}_{kj}\big),
\label{eq:margin}
\end{equation}
the total score the assignment would lose if the two identities were
exchanged, nonnegative by optimality and computed for pairs whose
cross-assignments both have spatial support. A pair with
$\Delta_{ab}<\tau_m\,{=}\,0.2$ is \emph{entangled}: the assignment's
score separates the chosen labeling from its swap by less than
$\tau_m$. Measured with the mechanism disabled (logging only; outputs
byte-identical), entanglement is rare and brief
(Sec.~\ref{sec:exp_arbitration}), which is precisely what makes
deferral affordable: positions remain unambiguous, only the
\emph{labels} are in doubt, and clean evidence surrounds a typically
one-frame ambiguity.

On first entanglement, \method{} opens an \emph{episode}: writing
$a,b$ for the pair's tracks, it snapshots their EMA appearance
prototypes $\phi_a, \phi_b$ \emph{before} any post-crossing update
can contaminate them. On each later frame where both match detections
with (unit-norm) embeddings $f_a, f_b$, the evidence over the set
$\mathcal{E}$ of such frames is
\begin{equation}
\ell \;=\; \sum_{t\in\mathcal{E}}
\big\langle f_a,\phi_a\big\rangle + \big\langle f_b,\phi_b\big\rangle
- \big\langle f_a,\phi_b\big\rangle - \big\langle f_b,\phi_a\big\rangle ;
\label{eq:llr}
\end{equation}
under a von Mises--Fisher observation model with shared
concentration, each term is (up to the concentration constant,
absorbed into the threshold) the per-frame log-likelihood ratio
against the swapped hypothesis. Persistently negative evidence says
each track now follows the \emph{other} identity. The episode closes
$W{=}30$ frames after its last entanglement (re-entanglement
extends the window and $\ell$ is never reset), or when a member
dies, which forfeits the swap. At closure, if
$\ell \le -\tau_\ell\,{=}\,{-}0.5$ with $|\mathcal{E}|\ge 4$,
the two live tracks exchange IDs: at most one corrective swap per
episode, so a wrong crossing decision costs at most the episode plus
$W$ frames of mislabeling, not the remainder of both tracks.
Everything is arithmetic on embeddings the tracker already computes
(${\approx}0.6$\,ms/frame). The swap also purifies the offline
prototypes, so deferral and the merge of Sec.~\ref{sec:merge}
compound (Sec.~\ref{sec:exp_arbitration}).

\subsection{The alternatives: same signal, other couplings}
\label{sec:alternatives}

To show that Eq.~\eqref{eq:adaptive} is the right use of $S$, we
implement the alternatives inside the same pipeline, each consuming
the same per-track $(\hat{\mathbf{z}}, S)$, and measure them on
identical inputs (Sec.~\ref{sec:exp_ladder}): a \emph{veto} that
rejects a pair whose Mahalanobis distance under $S$ exceeds $\tau_M$
(the DeepSORT use~\cite{wojke2017deepsort}; swept
$\tau_M{\in}\{9.49,\dots,100\}$, alone and stacked on the adaptive
expansion); a \emph{fixed} factor $e\equiv0.7$ for every track
(Deep-EIoU's initial value~\cite{huang2024deepeiou}); a
MeMoSORT-style binary \emph{switch} between two fixed factors on
normalized speed~\cite{wang2025memosort}; an \emph{anisotropic}
variant with separate $e_x,e_y$ from $\sigma_x/w$ and $\sigma_y/h$;
and a \emph{$\sigma$-DIoU} that replaces the dilated overlap by a
5-point unscented estimate of $\mathbb{E}[\mathrm{DIoU}]$ under
$\mathcal N(\hat{\mathbf{z}}_{xy}, S_{xy})$, optionally blended with
Eq.~\eqref{eq:adaptive} by $\beta$. Each mode is tuned at its own
matching threshold, a protocol frozen before the runs; full
definitions are in Appendix~\ref{sec:supp_alternatives}.

\subsection{Long horizon: appearance-only tracklet merge}
\label{sec:merge}

Sports broadcasts remove players from view for seconds: on
SportsMOT-val football, 64\% of identity switches follow gaps over
20 frames, 97\% of those full camera exits (median 111 frames), with
players well-detected at both gap edges: an out-of-view identity
failure, unreachable by in-view association geometry.

Our offline stage runs on the online output after linear
interpolation of gaps $\le$10 frames. Each tracklet carries an
appearance \emph{prototype} (re-normalized mean of its per-frame
embeddings); ordered pairs with temporal gap $\le 300$ link greedily
in descending prototype cosine order (threshold $0.85$), under a
union-find with each tail and head usable once.

The customary ingredient we \emph{remove} is the motion gate, which
extrapolates a tracklet's terminal velocity (a five-frame tail
estimate) across the gap and vetoes candidates landing more than
$\tau_v{=}2$ box diagonals from the next tracklet's start. Over a
hundred-frame gap under a panning camera that extrapolation carries
no information, so the veto must go and appearance must carry the
decision alone; empirically the gate, not the threshold or gap cap,
was the binding constraint (Sec.~\ref{sec:exp_merge}).


\section{Experiments}
\label{sec:experiments}

\begin{figure*}[t]
\centering
\includegraphics[width=0.86\textwidth]{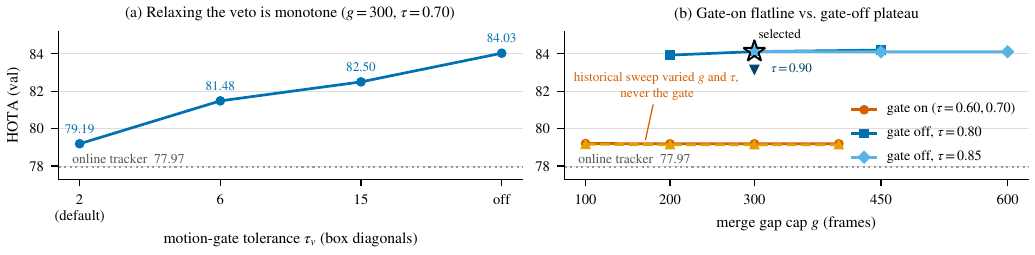}
\caption{\textbf{The motion gate is the binding constraint}
(SportsMOT val, pooled HOTA). (a) At $g{=}300$, relaxing the gate
tolerance is monotone: default 79.19 $\to$ removed 84.03. (b) Gate
on, sweeping gap cap and threshold is a flatline at ${\approx}79.2$;
gate off, a plateau (83.9--84.2), a robust optimum rather than a spike.
Star: selected configuration (best worst-sport HOTA).}
\label{fig:merge_grid}
\end{figure*}

\subsection{Setup}
\label{sec:exp_setup}

\paragraph{Datasets.} SportsMOT~\cite{cui2023sportsmot}: 45 val
sequences (15 each basketball, football, volleyball) for all tuning
and ablations; 150 test sequences evaluated exclusively through the
official evaluation server. DanceTrack~\cite{sun2022dancetrack} val
(25 sequences) serves as a transfer benchmark: non-linear motion
under a largely static camera. We report HOTA~\cite{luiten2021hota} with its detection
(DetA) and association (AssA) components, identity F1 (IDF1), multi-object
tracking accuracy (MOTA), and the number of identity switches (IDs).

\paragraph{Implementation.} Nothing in the system is trained. The
detector is YOLOX-X~\cite{ge2021yolox} with the public SportsMOT
weights from the MixSort model zoo~\cite{cui2023sportsmot}. Two
releases exist, and we keep them separate: every validation-set
experiment and ablation in this paper uses the checkpoint trained on
the \emph{train split only}, so no validation frame was seen by the
detector; the test-set submissions use the train$+$val checkpoint, the
release intended for test submissions. Validation-set DetA is
therefore not inflated by detector exposure;
appearance embeddings come from the sports-domain OSNet released with
Deep-EIoU~\cite{huang2024deepeiou} (DanceTrack rows use FastReID
SBS-S50~\cite{he2020fastreid}).
Association hyperparameters are given in
Secs.~\ref{sec:adaptive_eiou}--\ref{sec:merge}; the input-identical
runs change only detector-calibration thresholds
($\tau_H\,0.6{\to}0.5$, spawn $0.7{\to}0.6$, match $0.25{\to}0.2$).
All experiments run on a single RTX~3060 laptop GPU (6\,GB).
Detections, embeddings, and camera transforms are cached once, so
every ablation shares byte-identical inputs; an md5-verified
regression anchor guarantees the baseline configuration reproduces
exactly across the campaign.

\paragraph{Runtime.} With every mechanism active, association alone
runs at 204.8 frames per second (FPS); with live OSNet extraction the
tracker sustains ${\sim}65$\,FPS, and ${\sim}50$\,FPS including
JPEG decode, on the same 6\,GB laptop GPU (breakdown in
Appendix~\ref{sec:supp_transfer}). Table~\ref{tab:runtime_comparison} sets these numbers against the
throughput each competitor reports for itself; hardware differs per
row, so it bounds rather than ranks, but every trained SORT-family
entry reports under 30~FPS end to end and the mask-propagation entry
under 5.

\subsection{Comparison on SportsMOT test}
\label{sec:exp_sota}

\begin{table}[t]
\centering
\caption{\textbf{Comparison on the SportsMOT test set.}
  Competitor rows are from each paper's own results table (``--'':
  not reported); \method{} rows from the evaluation server.
  ``Tr.'': trains an association, motion, or appearance component.
  ``Off.'': offline tracklet-level (re-)association.}
\label{tab:test_comparison}
\resizebox{\columnwidth}{!}{
\begin{tabular}{lcccccccc}
\toprule
Method & Tr. & Off. & HOTA$\uparrow$ & DetA$\uparrow$ & AssA$\uparrow$ & MOTA$\uparrow$ & IDF1$\uparrow$ & IDs$\downarrow$ \\
\midrule
DiffMOT~\cite{lv2024diffmot} & \checkmark & -- & 76.2 & 89.3 & 65.1 & 97.1 & 76.1 & -- \\
McByte~\cite{stanczyk2025notrain} & -- & -- & 76.9 & -- & -- & 97.2 & 77.5 & -- \\
Deep-EIoU~\cite{huang2024deepeiou} & -- & -- & 77.2 & 88.2 & 67.7 & 96.3 & 79.8 & 2659 \\
SportMamba~\cite{khanna2025sportmamba} & \checkmark & -- & 77.3 & 89.5 & 66.8 & 96.9 & 77.7 & -- \\
DeepMoveSORT~\cite{adzemovic2024deepmovesort} & \checkmark & -- & 78.7 & 88.1 & 70.3 & 96.5 & 81.7 & -- \\
Deep HM-SORT~\cite{granhenriksen2024hmsort} & -- & -- & 80.1 & 88.3 & 72.7 & 96.6 & 85.2 & 1703 \\
CAMELTrack~\cite{somers2025cameltrack} & \checkmark & -- & 80.4 & 88.8 & 72.8 & 96.3 & 84.8 & -- \\
NOOUGAT~\cite{missaoui2025nooougat}$^{\P}$ & \checkmark & -- & 81.0 & -- & 74.3 & 96.0 & 85.3 & -- \\
GTA (Deep-EIoU)~\cite{sun2024gta}$^{\ast}$ & -- & \checkmark & 81.0 & 88.2 & 74.5 & 96.3 & 86.5 & 2737 \\
TDLP~\cite{adzemovic2025tdlp} & \checkmark & -- & 81.9 & 88.0 & 76.3 & 95.8 & 87.5 & -- \\
MeMoSORT~\cite{wang2025memosort} & \checkmark & -- & 82.1 & 89.3 & 75.6 & 97.0 & 86.4 & -- \\
\midrule
SAM 3-Deep-EIoU~\cite{holmberg2026selective}$^{\S}$ & -- & \checkmark & 87.2 & -- & 84.2 & 98.1 & 93.6 & -- \\
\midrule
\method{} (online $+$ gated merge, input-id.)$^\dagger$ & -- & \checkmark & 75.67 & 87.10 & 65.78 & 96.61 & 78.79 & 1833 \\
\method{} ($+$ gate removed, input-id.)$^\dagger$ & -- & \checkmark & 80.31 & 87.04 & 74.14 & -- & 86.92 & 1320 \\
\textbf{\method{} (full, ours)} & -- & \checkmark & \textbf{81.82} & \textbf{87.84} & \textbf{76.24} & \textbf{97.09} & \textbf{88.47} & \textbf{1414}$^\ddagger$ \\
\bottomrule
\end{tabular}
}
\par\smallskip
{\footnotesize
$^\dagger$Input-identical: Deep-EIoU's public detections and embeddings, without deferred commitment; row 1 keeps the merge's default motion gate (nearly inert), row 2 removes it (Sec.~\ref{sec:merge}); its MOTA was not recorded.\\
$^\ddagger$Fewest identity switches on the 69-entry leaderboard at tied-third MOTA (97.09); rank 6 in HOTA as of August 2026.\\
$^{\ast}$Offline global tracklet association on Deep-EIoU.\quad$^{\P}$Online mode; offline variant 85.6 HOTA / 83.0 AssA / 92.3 IDF1.\quad$^{\S}$Foundation tier: SAM~3 mask propagation with global track association.
\par}
\end{table}

\begin{table}[t]
\centering
\caption{\textbf{Reported throughput} in frames per second (FPS). Competitor numbers are
  each paper's own claim on its own hardware and benchmark, so the
  column is indicative, not a controlled comparison; ``assoc.''
  excludes detection and embedding extraction, ``full'' includes
  them. \method{} is measured on a 6\,GB laptop GPU.}
\label{tab:runtime_comparison}
\resizebox{\columnwidth}{!}{
\begin{tabular}{llll}
\toprule
Method & FPS & Scope & Hardware / benchmark (as reported) \\
\midrule
McByte~\cite{stanczyk2025notrain} & 3--5 & full & A100 \\
TDLP~\cite{adzemovic2025tdlp} & 2.0 (11.4 assoc.) & full & consumer GPU, SportsMOT \\
NOOUGAT (online)~\cite{missaoui2025nooougat} & 12 & assoc. & V100, DanceTrack val \\
CAMELTrack~\cite{somers2025cameltrack} & 13 & full & MOT17 \\
Deep HM-SORT~\cite{granhenriksen2024hmsort} & ${\approx}13$ & full & L40 (16\,GB VM) \\
Deep-EIoU~\cite{huang2024deepeiou} & ${\approx}14.6$ & full & RTX 4080 \\
DiffMOT~\cite{lv2024diffmot} & 22.7 & full & RTX 3090, DanceTrack \\
MeMoSORT~\cite{wang2025memosort} & 28.8 & full & unstated, DanceTrack val \\
SportMamba~\cite{khanna2025sportmamba} & ${\approx}30$ & full & RTX 4090 \\
\midrule
\textbf{\method{}} & \textbf{50.7} (64.8 excl.\ decode) & full & RTX 3060 Laptop, SportsMOT val \\
\method{} & 204.8 & assoc. & same \\
\bottomrule
\end{tabular}
}
\end{table}

Table~\ref{tab:test_comparison} places \method{} on the SportsMOT
test set. Two of our rows decompose the offline stage on the \emph{exact
public detections and embeddings} of
Deep-EIoU~\cite{huang2024deepeiou}; the third is the full tracker
on our own stack. With the merge still
motion-gated (nearly inert), the tracker scores $75.67$,
\emph{below} their published $77.20$: no in-view machinery reaches a
failure that is out of view. Removing the gate yields $80.31$
($+4.64$): the binding-constraint claim proven on the test server,
and $+3.1$ over Deep-EIoU on identical inputs before deferred
commitment is enabled. The
headline row ($81.82$) is the full tracker on our own stack,
rank~6 of 69 entries as of August 2026. The closest comparison is
GTA~\cite{sun2024gta} (81.0), the other training-free entry with
offline re-association, built on Deep-EIoU's stack and keeping the
spatio-temporal constraints our merge removes: on matched inputs we
trail it by $0.7$; on our own stack we pass it.
Among the SORT family of Table~\ref{tab:test_comparison}, the two
entries above \method{} both train their machinery, MeMoSORT
(82.1, an LSTM-gated Kalman filter)~\cite{wang2025memosort} and TDLP
(81.9, learned link prediction)~\cite{adzemovic2025tdlp}, and the
foundation-model tier (87.2) sits at a different compute class.
\method{} trains nothing. Read column-wise, the table says where the
result comes from: our DetA (87.84) is the lowest of every row that
reports it, our AssA (76.24) is second only to trained TDLP (76.3),
our IDF1 (88.47) is the highest below the foundation tier, and our
MOTA (97.09) trails only McByte (97.2) and DiffMOT (97.1).

\paragraph{Identity stability.} \method{} commits 1{,}414 identity
switches, the fewest of any entry on the 69-entry leaderboard as of
August 2026 (foundation-model leader included), at tied-third
MOTA ($97.09$), which rules out the trivial route to few switches,
tracking less; the count includes deferred commitment's own swaps.
With the 30th-best DetA on the board, \method{} ranks 6th in HOTA
and 4th in IDF1.

\paragraph{Shared-detection study.} On val, seven SORT-family
trackers run on identical cached detections (raw outputs): \method{}
77.71 leads BoT-SORT (76.49) and five others (75.01--71.75), best in
AssA, IDF1, and IDs (Table~\ref{tab:supp_shared_det} in the appendix).

\subsection{Comparison on DanceTrack test}
\label{sec:exp_dancetrack}

\begin{table}[t]
\centering
\caption{\textbf{Comparison on the DanceTrack test set.}
  Competitor rows from each paper's own table (``--'': not
  reported); ``Tr.''/``Off.'' as in Table~\ref{tab:test_comparison}.
  $^{\dagger}$: public DanceTrack YOLOX-X detections (ByteTrack
  release), the same as ours; $^{\ast}$: own detector with extra
  data. \method{} uses those detections and Deep OC-SORT's SBS-S50
  embeddings.}
\label{tab:dancetrack_comparison}
\resizebox{\columnwidth}{!}{
\begin{tabular}{lccccccc}
\toprule
Method & Tr. & Off. & HOTA$\uparrow$ & DetA$\uparrow$ & AssA$\uparrow$ & MOTA$\uparrow$ & IDF1$\uparrow$ \\
\midrule
SORT~\cite{bewley2016sort}$^{\dagger}$ & -- & -- & 47.9 & 72.0 & 31.2 & 91.8 & 50.8 \\
DeepSORT~\cite{wojke2017deepsort}$^{\dagger}$ & -- & -- & 45.6 & 71.0 & 29.7 & 87.8 & 47.9 \\
ByteTrack~\cite{zhang2022bytetrack}$^{\dagger}$ & -- & -- & 47.3 & 71.6 & 31.4 & 89.5 & 52.5 \\
OC-SORT~\cite{cao2023ocsort}$^{\dagger}$ & -- & \checkmark & 55.1 & 80.4 & 40.4 & 92.2 & 54.9 \\
Deep OC-SORT~\cite{maggiolino2023deep_ocsort}$^{\dagger}$ & -- & -- & 61.3 & 82.2 & 45.8 & 92.3 & 61.5 \\
DiffMOT~\cite{lv2024diffmot}$^{\dagger}$ & \checkmark & -- & 62.3 & 82.5 & 47.2 & 92.8 & 63.0 \\
DeepMoveSORT~\cite{adzemovic2024deepmovesort}$^{\dagger}$ & \checkmark & -- & 63.0 & 82.0 & 48.6 & 92.6 & 65.0 \\
NOOUGAT (online)~\cite{missaoui2025nooougat} & \checkmark & -- & 65.9 & -- & 54.9 & 88.9 & 70.6 \\
TrackTrack~\cite{shim2025tracktrack}$^{\ast}$ & -- & \checkmark & 66.5 & -- & 52.9 & 93.6 & 67.8 \\
McByte~\cite{stanczyk2025notrain} & -- & -- & 67.1 & -- & -- & 92.9 & 68.1 \\
MeMoSORT~\cite{wang2025memosort}$^{\ast}$ & \checkmark & -- & 67.9 & 85.0 & 54.3 & 93.4 & 68.0 \\
CAMELTrack~\cite{somers2025cameltrack}$^{\dagger}$ & \checkmark & -- & 69.3 & 81.8 & 58.9 & 91.4 & 74.9 \\
MOTRv2~\cite{zhang2023motrv2}$^{\dagger}$ & \checkmark & -- & 69.9 & 83.0 & 59.0 & 91.9 & 71.7 \\
TDLP~\cite{adzemovic2025tdlp}$^{\dagger}$ & \checkmark & -- & 70.1 & 82.6 & 59.6 & 91.8 & 75.8 \\
\midrule
\method{} (no deferred commitment)$^{\dagger}$ & -- & \checkmark & 61.1 & 80.6 & 46.5 & 92.5 & 65.1 \\
\textbf{\method{} (full, ours)}$^{\dagger}$ & -- & \checkmark & \textbf{62.1} & \textbf{80.3} & \textbf{48.2} & \textbf{92.4} & \textbf{66.5} \\
\bottomrule
\end{tabular}
}
\end{table}

Table~\ref{tab:dancetrack_comparison} reports the same association
configuration on DanceTrack test with every metric and against
published baselines. Among trackers that train nothing and run on
the public detections, \method{} is the strongest: $+0.8$ HOTA,
$+2.4$ AssA, and $+5.0$ IDF1 over Deep OC-SORT, which uses the same
detections and the same embedding model, so the difference is
association alone; it matches the trained motion models DiffMOT and
DeepMoveSORT within one HOTA point. The trained association and
end-to-end systems (MOTRv2, CAMELTrack, TDLP) sit $+7$ to $+8$ HOTA
above, and their advantage is entirely AssA ($59$--$60$ vs.\
$48.2$); DetA is detector-bound and in line with the public-detector
tier. DanceTrack has a static camera and no camera exits, so camera
compensation is neutral and the merge is worth $+1.9$ rather than
$+6$; deferred commitment ($+1.0$ test) is the mechanism it
exercises. We do not claim DanceTrack state of the art: the
configuration transfers without retuning, and the remaining gap is
association, not detection.

\subsection{Mechanism ladder: how to spend $S$}
\label{sec:exp_ladder}

\begin{table}[t]
\centering
\caption{\textbf{Association-mechanism ladder} on SportsMOT val
  (45 sequences, shared cached detections and embeddings): the same
  per-track covariance offered to every mechanism. The veto collapses
  association, fixed and switched expansion under-use the signal,
  continuous adaptivity recovers the full gap.}
\label{tab:mechanism_ladder}
\resizebox{\columnwidth}{!}{
\begin{tabular}{lcccc}
\toprule
Mechanism & HOTA$\uparrow$ & AssA$\uparrow$ & IDF1$\uparrow$ & IDSW$\downarrow$ \\
\midrule
Mahalanobis veto (best $\tau_M$)$^{\mathparagraph}$ & 76.53 & 67.34 & 79.88 & 547 \\
Fixed EIoU ($e{=}0.7$) & 77.04 & 68.20 & 80.52 & 512 \\
Velocity switch (best of 3)$^\ast$ & 77.24 & 68.54 & 80.73 & 506 \\
Pure $\sigma$-DIoU ($\beta{=}0$, best $\tau$)$^{\mathsection}$ & 77.46 & 68.84 & 80.70 & 530 \\
\textbf{Covariance-adaptive (ours)} & \textbf{77.97} & \textbf{69.78} & \textbf{81.48} & \textbf{514} \\
Anisotropic per-axis ($s{=}2$) & 77.97 & 69.78 & 81.49 & 513 \\
$\sigma$-blend ($\beta{=}0.3$, $\tau{=}0.15$)$^\ddagger$ & 78.03 & 69.89 & 81.48 & 516 \\
\bottomrule
\end{tabular}
}
\par\smallskip
{\footnotesize
$^{\mathparagraph}$Hard gate over fixed $e{=}0.7$ (covariance as veto only); best of $\tau_M\in\{9.49..100\}$ shown ($\tau_M{=}100$; monotone down to 71.26); stacked on adaptive: 70.82.\\
$^\ast$MeMoSORT-style two-level expansion; best of three variants shown (MeMoSORT-default thresholds; others: 77.08, 76.86).\\
$^{\mathsection}$Best $\beta{=}0$ threshold: expectation over the predictive distribution penalizes uncertainty (concentration bias).\\
$^\ddagger$Paired per-sequence vs.\ covariance-adaptive: mean $\Delta$HOTA $-0.18$ [95\% confidence interval (CI) $-0.91$, $+0.57$], sign-test $p{=}0.61$; statistically indistinguishable.
\par}
\end{table}

Table~\ref{tab:mechanism_ladder} is the paper's central experiment:
the same per-track $(\hat{\mathbf{z}}, S)$ offered to every
mechanism of Sec.~\ref{sec:alternatives} on identical inputs,
per-mode threshold tuning. Ordering: veto $\ll$ fixed $<$ switched
$<$ $\sigma$-pure $<$ \textbf{continuous} $\approx$ aniso
$\approx$ $\sigma$-blend.

\emph{Veto fails:} the Mahalanobis gate collapses association
(71.26 at $\chi^2_4(0.95){=}9.49$, switches doubled), recovers
monotonically but never fully ($-0.5$ at $\tau_M{=}100$ vs.\
gate-free 77.04), and stacked on the adaptive expansion is the worst
rung (70.82): a gate can only remove matches, and it removes the
ones the widened tolerance exists to keep.
\emph{Switching recovers a fifth:} the binary velocity switch peaks
at 77.24, recovering about a fifth of the 0.93 fixed$\to$adaptive
gap, its best variant the most conservative (aggressive variants:
77.08, 76.86).
\emph{Enrichment adds nothing:} anisotropic expansion is
byte-identical to the scalar on 41/45 sequences, the sigma-point
blend is statistically indistinguishable, and pure expected overlap
($\beta{=}0$) loses ${\approx}0.5$ even after retuning, since an
expectation penalizes uncertainty where expansion rewards it.

Together: the scalar $\max(\sigma_x/w,\sigma_y/h)$ is empirically a
sufficient statistic for association tolerance, and the covariance
must be spent as tolerance, not veto.

\paragraph{Scope.} The adaptive gain over fixed expansion ($+1.03$
mean per-sequence delta, 95\% confidence interval (CI)
$[+0.08,+2.13]$; pooled $+0.93$) concentrates in basketball
($+2.71$); the identity-switch improvement is broad-based (22W/7L,
$p{=}0.008$). Football is insensitive to every expansion mode, which
motivated the long-horizon analysis below. In the online build-up
($68.98 \to 77.97$), camera compensation ($+3.23$) and the IMM
filter ($+3.06$) dominate.

\subsection{Long horizon: the merge and its gate}
\label{sec:exp_merge}

Fig.~\ref{fig:merge_grid} quantifies the long-horizon claim on
SportsMOT val. The online tracker scores 77.97 HOTA (AssA 69.78,
IDSW 514); the merge with its customary motion gate ($g{=}100$,
$\tau{=}0.70$) reaches 79.20 (AssA 72.11, IDSW 436); the same merge
with the gate removed ($g{=}300$, $\tau{=}0.85$) reaches 84.12 (AssA
81.37, IDF1 91.72, IDSW 273). With the veto active the merge
plateaus at ${\approx}79.2$ regardless of gap cap or threshold
(Fig.~\ref{fig:merge_grid}b); relaxing it is monotone
(Fig.~\ref{fig:merge_grid}a), and with it removed the surface is a
flat plateau, a robust optimum rather than a spike. Fig.~\ref{fig:supp_qualitative} in the
appendix shows one instance, an 11.7-second camera exit that
only the appearance-only merge survives (the gated variant splits
the identity $3 \to 36$), and Table~\ref{tab:supp_merge_grid} the full grid. We select by best
\emph{worst-sport} HOTA on that plateau, not the pooled argmax. Football gains $+7.6$ over the online tracker;
appearance at cosine $\ge 0.85$ carries the decision even between
same-jersey teammates.

The test set arbitrates: the online and merged input-identical rows
of Table~\ref{tab:test_comparison} differ \emph{only} in this stage,
$+4.64$ on test against the $+4.92$ that removing the gate
delivered on val (pooled, gate-on to gate-off), a generalization tax
of ${\approx}0.3$. The identical merge configuration also transfers
to DanceTrack val under a \emph{different} embedding model (FastReID
SBS-S50): $+1.87$ HOTA at the same threshold optimum, harvesting
occlusion fragmentation rather than camera exits; and where fixed
expansion \emph{hurts} DanceTrack (56.53 vs.\ 56.72),
covariance-adaptive helps (56.86), since it reduces to plain IoU
where motion is confident (merge transfer: Table~\ref{tab:transfer} in
the appendix).

\subsection{Deferred commitment: measure first, then intervene}
\label{sec:exp_arbitration}

\begin{table}[t]
\centering
\caption{\textbf{Deferred identity commitment}, identical thresholds
  on both benchmarks. ``Online'': tracker with interpolation;
  ``chain'': $+$ appearance-only merge; test rows from the evaluation
  servers.}
\label{tab:arbitration}
\resizebox{\columnwidth}{!}{
\begin{tabular}{lcccccc}
\toprule
 & \multicolumn{3}{c}{SportsMOT} & \multicolumn{3}{c}{DanceTrack} \\
\cmidrule(lr){2-4}\cmidrule(lr){5-7}
HOTA & online & chain & \textbf{test} & online & chain & \textbf{test} \\
\midrule
without & 77.97 & 84.12 & 80.82 & 56.86 & 58.73 & 61.1 \\
\textbf{with} & \textbf{78.80} & \textbf{84.96} & \textbf{81.82} & \textbf{58.36} & \textbf{60.62} & \textbf{62.1} \\
$\Delta$ & +0.83 & +0.84 & \textbf{+1.00} & +1.50 & +1.89 & \textbf{+1.00} \\
\bottomrule
\end{tabular}
}
\end{table}

The mechanism of Sec.~\ref{sec:arbitration} was built
measurement-first: with intervention disabled, the margin read of
Eq.~\eqref{eq:margin} is logging-only (outputs remain
byte-identical) and counts the ambiguity each benchmark actually
holds: $640$ entangled episodes on DanceTrack val against $135$ on
SportsMOT val (25.6 vs.\ 3.0 per sequence), median duration one
frame. Table~\ref{tab:arbitration} then reports
the intervention under identical thresholds everywhere.

On DanceTrack val the online tracker gains $+1.50$ HOTA
($56.86\!\to\!58.36$), growing to $+1.89$ through the chain
($58.73\!\to\!60.62$): corrective swaps purify tracklet
prototypes, so the merge links better. On SportsMOT val the online gain is $+0.83$ pooled
($77.97\!\to\!78.80$; mean per-sequence delta $+0.69$, 95\% CI
$[+0.15,+1.32]$, paired bootstrap) and passes through the chain
undiminished ($84.12\!\to\!84.96$). Both test servers agree:
$+1.00$ on SportsMOT ($80.82\!\to\!81.82$, 95 swaps across 56 of 150
sequences) and $+1.00$ on DanceTrack ($61.1\!\to\!62.1$), each
against the byte-identical system without the mechanism.

Three finer-grained checks (numbers in Appendix~\ref{sec:supp_finegrained}): the gain
lives where crossings do (basketball $+1.68$; football, whose
failures are camera exits, $+0.18$); a per-swap ground-truth (GT)
audit on SportsMOT val finds 11 verified-correct swaps and none
verified-incorrect; and on Deep-EIoU's embeddings the same val
experiment measures $+1.06$, so the margin fires whatever the
embedding stack (val gains transfer at 53--120\% across our two test
runs).

Identity switches \emph{rise} under the mechanism (1{,}238 $\to$
1{,}414 on test) because each swap relabels both tracks (95 swaps,
$+176$ switches). The trade is deliberate: a bounded mislabeling buys
the remainder of both tracks, the association metrics price it
correctly (AssA $+1.86$, IDF1 $+1.30$), and the count remains the
lowest on the leaderboard as of August 2026.


\section{Conclusion}
\label{sec:conclusion}

We presented \method{}, a training-free tracker in which the
tracker's own uncertainty governs its decisions: the IMM covariance
sets a continuous per-track association tolerance, the assignment's
permutation margin defers near-ambiguous commitments until evidence
settles them, and where motion information has run out an
appearance-only merge replaces the motion-gated one.

\paragraph{Limitations.} Deferred commitment is pairwise and
single-swap, and its gain tracks the embedding stack ($53$--$120\%$
of the validation delta). The long-horizon stage is offline. Public
model-zoo detection caps HOTA. On DanceTrack, $62.1$ is well below
trained specialists; on SportsMOT, trained and foundation-model
systems~\cite{holmberg2026selective} keep a $0.1$--$5.4$ HOTA edge.

\iftoggle{wacvfinal}{%
\paragraph{Acknowledgments.}
Built on top of OC-SORT~\cite{cao2023ocsort}. Reference
implementations consulted: ByteTrack~\cite{zhang2022bytetrack},
BoT-SORT~\cite{aharon2022botsort},
Deep OC-SORT~\cite{maggiolino2023deep_ocsort}.
}{}

{
    \small
    \bibliographystyle{ieeenat_fullname}
    \bibliography{references}

@inproceedings{bewley2016sort,
  title     = {Simple Online and Realtime Tracking},
  author    = {Bewley, Alex and Ge, Zongyuan and Ott, Lionel and Ramos, Fabio and Upcroft, Ben},
  booktitle = {2016 IEEE International Conference on Image Processing (ICIP)},
  pages     = {3464--3468},
  year      = {2016},
  doi       = {10.1109/ICIP.2016.7533003}
}

@inproceedings{wojke2017deepsort,
  title     = {Simple Online and Realtime Tracking with a Deep Association Metric},
  author    = {Wojke, Nicolai and Bewley, Alex and Paulus, Dietrich},
  booktitle = {2017 IEEE International Conference on Image Processing (ICIP)},
  pages     = {3645--3649},
  year      = {2017},
  doi       = {10.1109/ICIP.2017.8296962}
}

@inproceedings{zhang2022bytetrack,
  title     = {{ByteTrack}: Multi-Object Tracking by Associating Every Detection Box},
  author    = {Zhang, Yifu and Sun, Peize and Jiang, Yi and Yu, Dongdong and Weng, Fucheng and Yuan, Zehuan and Luo, Ping and Liu, Wenyu and Wang, Xinggang},
  booktitle = {European Conference on Computer Vision (ECCV)},
  pages     = {1--21},
  year      = {2022},
  doi       = {10.1007/978-3-031-20047-2_1}
}

@inproceedings{cao2023ocsort,
  title     = {Observation-Centric {SORT}: Rethinking {SORT} for Robust Multi-Object Tracking},
  author    = {Cao, Jinkun and Pang, Jiangmiao and Weng, Xinshuo and Khirodkar, Rawal and Kitani, Kris},
  booktitle = {Proceedings of the IEEE/CVF Conference on Computer Vision and Pattern Recognition (CVPR)},
  pages     = {9686--9696},
  year      = {2023}
}

@article{aharon2022botsort,
  title   = {{BoT-SORT}: Robust Associations Multi-Pedestrian Tracking},
  author  = {Aharon, Nir and Orfaig, Roy and Bobrovsky, Ben-Zion},
  journal = {arXiv preprint arXiv:2206.14651},
  year    = {2022}
}

@inproceedings{maggiolino2023deep_ocsort,
  title     = {Deep {OC-SORT}: Multi-Pedestrian Tracking by Adaptive Re-Identification},
  author    = {Maggiolino, Gerard and Ahmad, Adnan and Cao, Jinkun and Kitani, Kris},
  booktitle = {2023 IEEE International Conference on Image Processing (ICIP)},
  pages     = {3025--3029},
  year      = {2023},
  doi       = {10.1109/ICIP49359.2023.10222576}
}

@inproceedings{sun2022dancetrack,
  title     = {{DanceTrack}: Multi-Object Tracking in Uniform Appearance and Diverse Motion},
  author    = {Sun, Peize and Cao, Jinkun and Jiang, Yi and Yuan, Zehuan and Bai, Song and Kitani, Kris and Luo, Ping},
  booktitle = {Proceedings of the IEEE/CVF Conference on Computer Vision and Pattern Recognition (CVPR)},
  pages     = {20993--21002},
  year      = {2022}
}

@article{luiten2021hota,
  title   = {{HOTA}: A Higher Order Metric for Evaluating Multi-Object Tracking},
  author  = {Luiten, Jonathon and O{\v{s}}ep, Aljo{\v{s}}a and Dendorfer, Patrick and Torr, Philip H. S. and Geiger, Andreas and Leal-Taix{\'e}, Laura and Leibe, Bastian},
  journal = {International Journal of Computer Vision},
  volume  = {129},
  number  = {2},
  pages   = {548--578},
  year    = {2021},
  doi     = {10.1007/s11263-020-01375-2}
}

@article{ge2021yolox,
  title   = {{YOLOX}: Exceeding {YOLO} Series in 2021},
  author  = {Ge, Zheng and Liu, Songtao and Wang, Feng and Li, Zeming and Sun, Jian},
  journal = {arXiv preprint arXiv:2107.08430},
  year    = {2021}
}

@inproceedings{he2020fastreid,
  title     = {{FastReID}: A Pytorch Toolbox for General Instance Re-identification},
  author    = {He, Lingxiao and Liao, Xingyu and Liu, Wu and Liu, Xinchen and Cheng, Peng and Mei, Tao},
  booktitle = {Proceedings of the 31st ACM International Conference on Multimedia},
  pages     = {9664--9667},
  year      = {2023},
  doi       = {10.1145/3581783.3613460}
}

@article{blom1988interacting,
  title   = {The Interacting Multiple Model Algorithm for Systems with {M}arkovian Switching Coefficients},
  author  = {Blom, Henk A. P. and Bar-Shalom, Yaakov},
  journal = {IEEE Transactions on Automatic Control},
  volume  = {33},
  number  = {8},
  pages   = {780--783},
  year    = {1988},
  doi     = {10.1109/9.1299}
}

@book{barshalom2001estimation,
  title     = {Estimation with Applications to Tracking and Navigation: Theory Algorithms and Software},
  author    = {Bar-Shalom, Yaakov and Li, X. Rong and Kirubarajan, Thiagalingam},
  publisher = {John Wiley \& Sons},
  year      = {2001},
  isbn      = {978-0-471-41655-5},
  doi       = {10.1002/0471221279}
}

@article{mazor1998interacting,
  title   = {Interacting Multiple Model Methods in Target Tracking: A Survey},
  author  = {Mazor, E. and Averbuch, A. and Bar-Shalom, Y. and Dayan, J.},
  journal = {IEEE Transactions on Aerospace and Electronic Systems},
  volume  = {34},
  number  = {1},
  pages   = {103--123},
  year    = {1998},
  doi     = {10.1109/7.640267}
}

@inproceedings{yi2024ucmctrack,
  title     = {{UCMCTrack}: Multi-Object Tracking with Uniform Camera Motion Compensation},
  author    = {Yi, Kefu and Luo, Kai and Luo, Xiaolei and Huang, Jiangui and Wu, Hao and Hu, Rongdong and Hao, Wei},
  booktitle = {Proceedings of the AAAI Conference on Artificial Intelligence},
  volume    = {38},
  number    = {7},
  pages     = {6702--6710},
  year      = {2024},
  doi       = {10.1609/aaai.v38i7.28493}
}

@article{du2023strongsort,
  title   = {{StrongSORT}: Make {DeepSORT} Great Again},
  author  = {Du, Yunhao and Zhao, Zhicheng and Song, Yang and Zhao, Yanyun and Su, Fei and Gong, Tao and Meng, Hongying},
  journal = {IEEE Transactions on Multimedia},
  volume  = {25},
  pages   = {8725--8737},
  year    = {2023},
  doi     = {10.1109/TMM.2023.3240881}
}

@inproceedings{cui2023sportsmot,
  title     = {{SportsMOT}: A Large Multi-Object Tracking Dataset in Multiple Sports Scenes},
  author    = {Cui, Yutao and Zeng, Chenkai and Zhao, Xiaoyu and Yang, Yichun and Wu, Gangshan and Wang, Limin},
  booktitle = {Proceedings of the IEEE/CVF International Conference on Computer Vision (ICCV)},
  pages     = {9887--9897},
  year      = {2023},
  doi       = {10.1109/ICCV51070.2023.00910}
}

@inproceedings{yang2024hybridsort,
  title     = {Hybrid-{SORT}: Weak Cues Matter for Online Multi-Object Tracking},
  author    = {Yang, Mingzhan and Han, Guangxin and Yan, Bin and Zhang, Wenhua and Qi, Jinqing and Lu, Huchuan and Wang, Dong},
  booktitle = {Proceedings of the AAAI Conference on Artificial Intelligence},
  volume    = {38},
  number    = {7},
  pages     = {6504--6512},
  year      = {2024},
  doi       = {10.1609/aaai.v38i7.28471}
}

@inproceedings{shim2025tracktrack,
  title     = {Focusing on Tracks for Online Multi-Object Tracking},
  author    = {Shim, Kyujin and Ko, Kangwook and Yang, Yujin and Kim, Changick},
  booktitle = {Proceedings of the IEEE/CVF Conference on Computer Vision and Pattern Recognition (CVPR)},
  pages     = {11687--11696},
  year      = {2025},
  doi       = {10.1109/CVPR52734.2025.01091}
}

@inproceedings{yang2023cbiou,
  title     = {Hard to Track Objects with Irregular Motions and Similar Appearances? Make It Easier by Buffering the Matching Space},
  author    = {Yang, Fan and Odashima, Shigeyuki and Masui, Shoichi and Jiang, Shan},
  booktitle = {Proceedings of the IEEE/CVF Winter Conference on Applications of Computer Vision (WACV)},
  pages     = {4788--4797},
  year      = {2023},
  doi       = {10.1109/WACV56688.2023.00478}
}

@inproceedings{huang2024deepeiou,
  title     = {Iterative Scale-Up {ExpansionIoU} and Deep Features Association for Multi-Object Tracking in Sports},
  author    = {Huang, Hsiang-Wei and Yang, Cheng-Yen and Sun, Jiacheng and Kim, Pyong-Kun and Kim, Kwang-Ju and Lee, Kyoungoh and Huang, Chung-I and Hwang, Jenq-Neng},
  booktitle = {Proceedings of the IEEE/CVF Winter Conference on Applications of Computer Vision (WACV) Workshops},
  pages     = {163--172},
  year      = {2024},
  doi       = {10.1109/WACVW60836.2024.00024}
}

@article{granhenriksen2024hmsort,
  title   = {Deep {HM-SORT}: Enhancing Multi-Object Tracking in Sports with Deep Features, Harmonic Mean, and Expansion {IOU}},
  author  = {Gran-Henriksen, Matias and Lindgaard, Hans Andreas and Kiss, Gabriel and Lindseth, Frank},
  journal = {arXiv preprint arXiv:2406.12081},
  year    = {2024}
}

@article{wang2025memosort,
  title   = {{MeMoSORT}: Memory-Assisted Filtering and Motion-Adaptive Association Metric for Multi-Person Tracking},
  author  = {Wang, Yingjie and Wang, Zhixing and Zheng, Le and Liu, Tianxiao and Li, Roujing and Hu, Xueyao},
  journal = {arXiv preprint arXiv:2508.09796},
  year    = {2025}
}

@inproceedings{solanocarrillo2024utrack,
  title     = {{UTrack}: Multi-Object Tracking with Uncertain Detections},
  author    = {Solano-Carrillo, Edgardo and Sattler, Felix and Alex, Antje and Klein, Alexander and Pereira Costa, Bruno and Bueno Rodriguez, Angel and Stoppe, Jannis},
  booktitle = {Computer Vision -- ECCV 2024 Workshops},
  series    = {Lecture Notes in Computer Science},
  volume    = {15639},
  pages     = {219--236},
  publisher = {Springer},
  year      = {2025},
  doi       = {10.1007/978-3-031-91585-7_14}
}

@article{llerena2021probiou,
  title   = {Gaussian Bounding Boxes and Probabilistic Intersection-over-Union for Object Detection},
  author  = {Llerena, Jeffri M. and Zeni, Luis Felipe and Kristen, Lucas N. and Jung, Claudio},
  journal = {arXiv preprint arXiv:2106.06072},
  year    = {2021}
}

@inproceedings{wang2020jiou,
  title     = {Inferring Spatial Uncertainty in Object Detection},
  author    = {Wang, Zining and Feng, Di and Zhou, Yiyang and Rosenbaum, Lars and Timm, Fabian and Dietmayer, Klaus and Tomizuka, Masayoshi and Zhan, Wei},
  booktitle = {Proceedings of the IEEE/RSJ International Conference on Intelligent Robots and Systems (IROS)},
  pages     = {5792--5799},
  year      = {2020},
  doi       = {10.1109/IROS45743.2020.9340798}
}

@article{kirsten2023cell,
  title   = {Cell Tracking-by-Detection Using Elliptical Bounding Boxes},
  author  = {Kirsten, Lucas N. and Jung, Cl{\'a}udio R.},
  journal = {Journal of Visual Communication and Image Representation},
  volume  = {108},
  pages   = {104425},
  year    = {2025},
  doi     = {10.1016/j.jvcir.2025.104425}
}

@inproceedings{qin2023motiontrack,
  title     = {{MotionTrack}: Learning Robust Short-Term and Long-Term Motions for Multi-Object Tracking},
  author    = {Qin, Zheng and Zhou, Sanping and Wang, Le and Duan, Jinghai and Hua, Gang and Tang, Wei},
  booktitle = {Proceedings of the IEEE/CVF Conference on Computer Vision and Pattern Recognition (CVPR)},
  pages     = {17939--17948},
  year      = {2023},
  doi       = {10.1109/CVPR52729.2023.01720}
}

@inproceedings{lv2024diffmot,
  title     = {{DiffMOT}: A Real-time Diffusion-based Multiple Object Tracker with Non-linear Prediction},
  author    = {Lv, Weiyi and Huang, Yuhang and Zhang, Ning and Lin, Ruei-Sung and Han, Mei and Zeng, Dan},
  booktitle = {Proceedings of the IEEE/CVF Conference on Computer Vision and Pattern Recognition (CVPR)},
  pages     = {19321--19330},
  year      = {2024},
  doi       = {10.1109/CVPR52733.2024.01828}
}

@inproceedings{huang2024mambamot,
  title     = {{MambaMOT}: State-Space Model as Motion Predictor for Multi-Object Tracking},
  author    = {Huang, Hsiang-Wei and Yang, Cheng-Yen and Chai, Wenhao and Jiang, Zhongyu and Hwang, Jenq-Neng},
  booktitle = {Proceedings of the IEEE International Conference on Acoustics, Speech and Signal Processing (ICASSP)},
  pages     = {1--5},
  year      = {2025},
  doi       = {10.1109/ICASSP49660.2025.10890199}
}

@article{hu2024trackssm,
  title   = {{TrackSSM}: A General Motion Predictor Using a State Space Model},
  author  = {Hu, Bin and Luo, Run and Liu, Zelin and Wang, Cheng and Liu, Wenyu},
  journal = {Science China Information Sciences},
  volume  = {69},
  number  = {5},
  pages   = {152105},
  year    = {2026},
  doi     = {10.1007/s11432-024-4849-2}
}

@inproceedings{khanna2025sportmamba,
  title     = {{SportMamba}: Adaptive Non-Linear Multi-Object Tracking with State Space Models for Team Sports},
  author    = {Khanna, Dheeraj and Bright, Jerrin and Chen, Yuhao and Zelek, John S.},
  booktitle = {Proceedings of the IEEE/CVF Conference on Computer Vision and Pattern Recognition (CVPR) Workshops},
  pages     = {6133--6143},
  year      = {2025},
  doi       = {10.1109/CVPRW67362.2025.00611}
}

@article{adzemovic2024movesort,
  title   = {Beyond {Kalman} Filters: Deep Learning-Based Filters for Improved Object Tracking},
  author  = {Ad{\v{z}}emovi{\'c}, Momir and Tadi{\'c}, Predrag and Petrovi{\'c}, Andrija and Nikoli{\'c}, Mladen},
  journal = {Machine Vision and Applications},
  volume  = {36},
  number  = {1},
  pages   = {20},
  year    = {2025},
  doi     = {10.1007/s00138-024-01644-x}
}

@article{adzemovic2024deepmovesort,
  title   = {Engineering an Efficient Object Tracker for Non-Linear Motion},
  author  = {Ad{\v{z}}emovi{\'c}, Momir and Tadi{\'c}, Predrag and Petrovi{\'c}, Andrija and Nikoli{\'c}, Mladen},
  journal = {arXiv preprint arXiv:2407.00738},
  year    = {2024}
}

@article{somers2025cameltrack,
  title   = {{CAMELTrack}: Context-Aware Multi-cue {ExpLoitation} for Online Multi-Object Tracking},
  author  = {Somers, Vladimir and Standaert, Baptiste and Joos, Victor and Alahi, Alexandre and De Vleeschouwer, Christophe},
  journal = {arXiv preprint arXiv:2505.01257},
  year    = {2025}
}

@inproceedings{zeng2022motr,
  title     = {{MOTR}: End-to-End Multiple-Object Tracking with Transformer},
  author    = {Zeng, Fangao and Dong, Bin and Zhang, Yuang and Wang, Tiancai and Zhang, Xiangyu and Wei, Yichen},
  booktitle = {Computer Vision -- ECCV 2022},
  series    = {Lecture Notes in Computer Science},
  volume    = {13687},
  pages     = {659--675},
  publisher = {Springer},
  year      = {2022},
  doi       = {10.1007/978-3-031-19812-0_38}
}

@inproceedings{gao2025motip,
  title     = {Multiple Object Tracking as {ID} Prediction},
  author    = {Gao, Ruopeng and Qi, Ji and Wang, Limin},
  booktitle = {Proceedings of the IEEE/CVF Conference on Computer Vision and Pattern Recognition (CVPR)},
  pages     = {27883--27893},
  year      = {2025},
  doi       = {10.1109/CVPR52734.2025.02596}
}

@article{barshalom1975pdaf,
  title   = {Tracking in a cluttered environment with probabilistic data association},
  author  = {Bar-Shalom, Yaakov and Tse, Edison},
  journal = {Automatica},
  volume  = {11},
  number  = {5},
  pages   = {451--460},
  year    = {1975},
  doi     = {10.1016/0005-1098(75)90021-7}
}

@article{fortmann1983jpdaf,
  title   = {Sonar tracking of multiple targets using joint probabilistic data association},
  author  = {Fortmann, Thomas and Bar-Shalom, Yaakov and Scheffe, Molly},
  journal = {IEEE Journal of Oceanic Engineering},
  volume  = {8},
  number  = {3},
  pages   = {173--184},
  year    = {1983},
  doi     = {10.1109/JOE.1983.1145560}
}

@inproceedings{sun2024gta,
  title     = {{GTA}: Global Tracklet Association for Multi-Object Tracking in Sports},
  author    = {Sun, Jiacheng and Huang, Hsiang-Wei and Yang, Cheng-Yen and Jiang, Zhongyu and Hwang, Jenq-Neng},
  booktitle = {Asian Conference on Computer Vision (ACCV) Workshops},
  year      = {2024}
}

@inproceedings{stanczyk2025notrain,
  title     = {No Train Yet Gain: Towards Generic Multi-Object Tracking in Sports and Beyond},
  author    = {Stanczyk, Tomasz and Yoon, Seongro and Bremond, Francois},
  booktitle = {Proceedings of the IEEE/CVF Conference on Computer Vision and Pattern Recognition (CVPR) Workshops},
  year      = {2025}
}

@article{missaoui2025nooougat,
  title   = {{NOOUGAT}: Towards Unified Online and Offline Multi-Object Tracking},
  author  = {Missaoui, Benjamin and Cetintas, Orcun and Bras{\'o}, Guillem and Meinhardt, Tim and Leal-Taix{\'e}, Laura},
  journal = {arXiv preprint arXiv:2509.02111},
  year    = {2025}
}

@article{adzemovic2025tdlp,
  title   = {Learning Association via Track-Detection Matching for Multi-Object Tracking},
  author  = {Ad{\v{z}}emovi{\'c}, Momir},
  journal = {arXiv preprint arXiv:2512.22105},
  year    = {2025}
}

@article{holmberg2026selective,
  title   = {Selective Mask Propagation for Multi-Object Tracking},
  author  = {Holmberg, Alexander},
  journal = {arXiv preprint arXiv:2606.13033},
  year    = {2026}
}

@article{reid1979mht,
  author  = {Reid, Donald B.},
  title   = {An Algorithm for Tracking Multiple Targets},
  journal = {IEEE Transactions on Automatic Control},
  volume  = {24},
  number  = {6},
  pages   = {843--854},
  year    = {1979},
  doi     = {10.1109/TAC.1979.1102177}
}

@inproceedings{kim2015mht,
  author    = {Kim, Chanho and Li, Fuxin and Ciptadi, Arridhana and Rehg, James M.},
  title     = {Multiple Hypothesis Tracking Revisited},
  booktitle = {Proceedings of the IEEE International Conference on Computer Vision (ICCV)},
  pages     = {4696--4704},
  year      = {2015},
  doi       = {10.1109/ICCV.2015.533}
}

@inproceedings{rezatofighi2015jpda,
  author    = {Rezatofighi, Seyed Hamid and Milan, Anton and Zhang, Zhen and Shi, Qinfeng and Dick, Anthony and Reid, Ian},
  title     = {Joint Probabilistic Data Association Revisited},
  booktitle = {Proceedings of the IEEE International Conference on Computer Vision (ICCV)},
  pages     = {3047--3055},
  year      = {2015},
  doi       = {10.1109/ICCV.2015.349}
}

@inproceedings{choi2015nomt,
  author    = {Choi, Wongun},
  title     = {Near-Online Multi-target Tracking with Aggregated Local Flow Descriptor},
  booktitle = {Proceedings of the IEEE International Conference on Computer Vision (ICCV)},
  pages     = {3029--3037},
  year      = {2015},
  doi       = {10.1109/ICCV.2015.347}
}

@inproceedings{zhang2023motrv2,
  author    = {Yuang Zhang and Tiancai Wang and Xiangyu Zhang},
  title     = {{MOTRv2}: Bootstrapping End-to-End Multi-Object Tracking by Pretrained Object Detectors},
  booktitle = {Proceedings of the IEEE/CVF Conference on Computer Vision and Pattern Recognition (CVPR)},
  pages     = {22056--22065},
  year      = {2023}
}
}

\clearpage
\appendix

\section{Deferred commitment: fine-grained checks (full numbers)}
\label{sec:supp_finegrained}
Per-sport arbitration delta on SportsMOT val (interp): basketball
$+1.68$, volleyball $+0.21$, football $+0.18$; corrective swaps
12/11/2 respectively: football's failures are camera exits, and
deferral correctly leaves them alone. Per-swap GT audit
(episode-anchored windows): SportsMOT 11 verified-correct / 0
verified-incorrect / 14 occlusion-undecidable of 25 swaps; DanceTrack
11 / 24 / 53 of 88, where crossing density defeats pre/post
attribution and the sequence-level paired deltas ($+1.50$ online,
$+1.00$ test) carry the verdict. Embedding dependence: on Deep-EIoU's
embeddings the same val experiment measures $+1.06$ pooled (144
entangled episodes, 18 swaps).

\section{The alternatives: full definitions}
\label{sec:supp_alternatives}

The claim that Eq.~\eqref{eq:adaptive} is the \emph{right} use of
$S$ is defended by implementing the alternatives inside the same
pipeline and measuring them under identical inputs
(Sec.~\ref{sec:exp_ladder}). Each consumes exactly the same per-track
$(\hat{\mathbf{z}}, S)$:

\noindent\textbf{Veto (Mahalanobis gate).} Reject pair $(n,m)$ if
its Mahalanobis distance under $S_m$ exceeds $\tau_M$, the
classical DeepSORT-style use~\cite{wojke2017deepsort}, swept over
$\tau_M \in \{9.49, 16, 30, 60, 100\}$ with fixed $e{=}0.7$
(covariance consumed \emph{only} as veto), plus a variant stacked on
the adaptive expansion.

\noindent\textbf{Fixed expansion.} $e_i \equiv 0.7$ for every
track, Deep-EIoU's initial factor~\cite{huang2024deepeiou}.

\noindent\textbf{Binary velocity switch.} MeMoSORT-style
\cite{wang2025memosort}: normalized speed thresholded at a
percentile selects between two fixed factors; we evaluate their
default calibration and two more aggressive variants.

\noindent\textbf{Anisotropic expansion.} Per-axis factors from
$\sigma_x/w$ and $\sigma_y/h$ separately; does directional shape
help?

\noindent\textbf{Sigma-point expected overlap ($\sigma$-DIoU).}
The full distributional treatment: an unscented 5-point approximation
of the expected DIoU under $\mathcal N(\hat{\mathbf{z}}_{xy},
S_{xy})$ (this variant does see $\Sigma_{xy}$), optionally blended
with Eq.~\eqref{eq:adaptive} by a factor $\beta$. Each mode is tuned
at its own matching threshold, a protocol frozen before the runs.

\begin{figure}[t]
\centering
\includegraphics[width=\columnwidth]{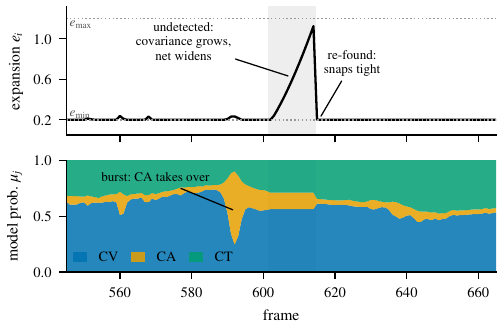}
\caption{\textbf{The mechanism live} (SportsMOT val basketball, one
track; recorder byte-identical to the ablation anchor). Top: $e_i$
(Eq.~\eqref{eq:adaptive}) hugs $e_{\min}$ under confident tracking,
bumps at maneuver onsets, grows toward $e_{\max}$ across a 12-frame
miss (gray), and snaps tight on re-capture. Bottom: IMM model
probabilities; constant-acceleration takes over during the frame-590
burst: disagreement widens $S$, and the net widens exactly then.}
\label{fig:supp_mechanism}
\end{figure}

\begin{figure*}[t]
\centering
\includegraphics[width=0.86\textwidth]{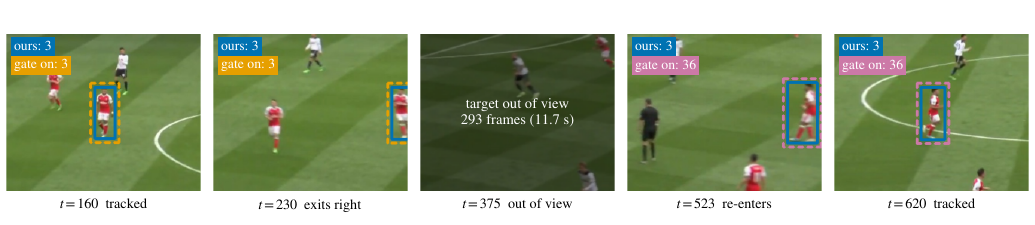}
\caption{\textbf{A camera exit, survived} (SportsMOT val football).
The target (solid blue) leaves the frame at $t{=}230$ and re-enters
$293$ frames ($11.7$\,s) later. The appearance-only merge holds one
identity across the gap (GT-verified); with the motion veto active
(dashed), the terminal-velocity extrapolation lands nowhere near the
re-entry and the identity splits $3 \to 36$ (amber $\to$ purple).}
\label{fig:supp_qualitative}
\end{figure*}

\section{Video}
\label{sec:supp_video}

The ancillary video file \texttt{camera\_exit.mp4} (listed under
``Ancillary files'' on the arXiv abstract page) shows the
qualitative episode of Fig.~\ref{fig:supp_qualitative} in motion: SportsMOT val
football sequence \texttt{v\_ITo3sCnpw\_k\_c012}, rendered
stacked with the appearance-only merge (top, ours: the target
keeps identity~3 across the 11.7-second camera exit) and with the
customary motion-extrapolation veto active (bottom: the identity
splits $3 \to 36$ at re-entry). Both runs share identical detections,
embeddings, and camera transforms; the only difference is the merge
gate.

\paragraph{Exemplar selection protocol.} The episode was selected
programmatically, not by hand: over all football val sequences we
enumerated merged tracks whose largest intra-track gap exceeds 60
frames, required the motion-gated output to assign different
identities to the two gap edges (box IoU $> 0.7$ against the merged
track on both sides), and verified against ground truth that both
edges belong to the same GT identity (IoU 0.88 / 0.93 here). The
selected episode is the longest GT-verified such gap (293 frames).

\section{Full mechanism-ladder measurements}
\label{sec:supp_ladder}

Table~\ref{tab:supp_ladder_full} lists every rung of the three
mechanism campaigns behind Table~\ref{tab:mechanism_ladder}, including all
Mahalanobis-gate thresholds (monotone recovery from 71.26 at
$\chi^2_4(0.95)$ to 76.53 at $\tau_M{=}100$, never reaching the
gate-free 77.04), all velocity-switch calibrations, and the full
$\sigma$-DIoU $\beta \times \tau$ sweep. The veto rungs were
re-measured in the final-configuration harness
(\texttt{ablation\_mahal\_final}); its rung 00 reproduces the 77.97
anchor exactly, so all rows are same-harness comparable. Historical
Phase-A gate numbers (measured against an earlier 76.7-HOTA
configuration) are retained in the repository but are deliberately
not tabulated here, as they are not comparable to the ladder anchor.

\begin{table}[t]
\centering
\caption{\textbf{Every measured rung} of the mechanism
  campaigns on SportsMOT val (shared cached inputs; each
  campaign's rung 00 reproduces the 77.97 anchor exactly).
  Velocity-switch calibration: MeMoSORT defaults
  $\theta{=}0.1172$, $p{=}0.5/0.6$; ``strong fast''
  $p_{\text{fast}}{=}1.0$; ``low thresh''
  $\theta{=}0.05$, $p_{\text{fast}}{=}1.0$.}
\label{tab:supp_ladder_full}
\resizebox{\columnwidth}{!}{
\begin{tabular}{lcccc}
\toprule
Rung & HOTA$\uparrow$ & AssA$\uparrow$ & IDF1$\uparrow$ & IDSW$\downarrow$ \\
\midrule
\multicolumn{5}{l}{\emph{Mode ladder (\texttt{ablation\_eiou\_modes})}} \\
\texttt{00\_repro\_adaptive\_s2} & 77.97 & 69.78 & 81.48 & 514 \\
\texttt{01\_fixed\_e07} & 77.04 & 68.20 & 80.52 & 512 \\
\texttt{02\_vswitch\_memo\_defaults} & 77.24 & 68.54 & 80.73 & 506 \\
\texttt{03\_vswitch\_strong\_fast} & 77.08 & 68.28 & 80.51 & 515 \\
\texttt{04\_vswitch\_low\_thresh} & 76.86 & 67.95 & 80.35 & 529 \\
\texttt{05\_aniso\_s2} & 77.97 & 69.78 & 81.49 & 513 \\
\texttt{06\_aniso\_s3} & 77.71 & 69.39 & 81.20 & 510 \\
\texttt{07\_aniso\_s2\_cap15} & 77.98 & 69.78 & 81.49 & 511 \\
\midrule
\multicolumn{5}{l}{\emph{$\sigma$-DIoU sweep (\texttt{ablation\_sigma\_iou})}} \\
\texttt{10\_sigma\_b0\_t25} & 77.29 & 68.46 & 80.12 & 600 \\
\texttt{11\_sigma\_b0\_t15} & 77.40 & 68.67 & 80.45 & 561 \\
\texttt{12\_sigma\_b0\_t05} & 77.46 & 68.84 & 80.70 & 530 \\
\texttt{13\_sigma\_b3\_t25} & 77.90 & 69.60 & 81.06 & 566 \\
\texttt{14\_sigma\_b3\_t15} & 78.03 & 69.89 & 81.48 & 516 \\
\texttt{15\_sigma\_b3\_t05} & 77.93 & 69.73 & 81.41 & 509 \\
\texttt{16\_sigma\_b5\_t25} & 77.83 & 69.54 & 81.26 & 547 \\
\texttt{17\_sigma\_b5\_t15} & 77.86 & 69.64 & 81.38 & 514 \\
\texttt{18\_sigma\_b7\_t15} & 77.49 & 69.02 & 80.97 & 515 \\
\midrule
\multicolumn{5}{l}{\emph{Mahalanobis veto (\texttt{ablation\_mahal\_final})}} \\
\texttt{00\_repro\_adaptive\_s2} & 77.97 & 69.78 & 81.48 & 514 \\
\texttt{01\_veto\_t949\_fixed} & 71.26 & 59.02 & 72.31 & 1127 \\
\texttt{02\_veto\_t16\_fixed} & 72.58 & 60.94 & 74.40 & 845 \\
\texttt{03\_veto\_t30\_fixed} & 74.99 & 64.77 & 77.52 & 658 \\
\texttt{04\_veto\_t60\_fixed} & 76.01 & 66.49 & 79.06 & 579 \\
\texttt{05\_veto\_t100\_fixed} & 76.53 & 67.34 & 79.88 & 547 \\
\texttt{06\_veto\_t949\_adaptive} & 70.82 & 58.27 & 71.81 & 1150 \\
\bottomrule
\end{tabular}
}
\end{table}

\section{Full merge grid}
\label{sec:supp_merge}

Table~\ref{tab:supp_merge_grid} lists every cell of the offline-merge
grid summarized in Fig.~\ref{fig:merge_grid}: the gate-on flatline
($79.14$--$79.20$ across every gap cap and threshold), the monotone
tolerance dose-response (tol $2 \to 6 \to 15 \to$ off), and the
gate-off plateau, including the pooled argmax (g450/off/0.80, 84.19)
that we deliberately do not select; the selected configuration
(g300/off/0.85) maximizes worst-sport HOTA instead.

\begin{table}[t]
\centering
\caption{\textbf{The full offline merge grid} on SportsMOT
  val (pooled). ``tol'': motion-gate tolerance in box
  diagonals (2.0 = customary default; off = removed). The
  pooled argmax is g450/off/0.80 (84.19); we select
  g300/off/0.85 by best worst-sport HOTA.}
\label{tab:supp_merge_grid}
\resizebox{\columnwidth}{!}{
\begin{tabular}{lccccc}
\toprule
Run & tol & HOTA$\uparrow$ & AssA$\uparrow$ & IDF1$\uparrow$ & IDSW$\downarrow$ \\
\midrule
\texttt{merge\_g100\_t06} & 2.0 & 79.16 & 72.04 & 83.60 & 434 \\
\texttt{merge\_g100\_t07} & 2.0 & 79.20 & 72.11 & 83.65 & 436 \\
\texttt{merge\_g200\_mtoff\_t08} & off & 83.93 & 80.96 & 91.49 & 258 \\
\texttt{merge\_g200\_t06} & 2.0 & 79.14 & 72.00 & 83.59 & 434 \\
\texttt{merge\_g200\_t07} & 2.0 & 79.19 & 72.09 & 83.64 & 436 \\
\texttt{merge\_g300\_mt1000000\_t07} & $10^6$ & 84.03 & 81.23 & 91.83 & 248 \\
\texttt{merge\_g300\_mt1000000\_t08} & $10^6$ & 84.11 & 81.35 & 91.84 & 260 \\
\texttt{merge\_g300\_mt15\_t07} & 15 & 82.50 & 78.27 & 89.51 & 287 \\
\texttt{merge\_g300\_mt15\_t08} & 15 & 82.52 & 78.26 & 89.31 & 302 \\
\texttt{merge\_g300\_mt6\_t07} & 6 & 81.48 & 76.32 & 87.41 & 357 \\
\texttt{merge\_g300\_mt6\_t08} & 6 & 81.49 & 76.30 & 87.26 & 369 \\
\texttt{merge\_g300\_mtoff\_t085} & off & 84.12 & 81.37 & 91.72 & 273 \\
\texttt{merge\_g300\_mtoff\_t08\_nosplit} & off & 84.11 & 81.35 & 91.84 & 260 \\
\texttt{merge\_g300\_mtoff\_t09} & off & 83.17 & 79.42 & 90.00 & 348 \\
\texttt{merge\_g300\_t06} & 2.0 & 79.14 & 72.00 & 83.59 & 434 \\
\texttt{merge\_g300\_t07} & 2.0 & 79.19 & 72.09 & 83.64 & 436 \\
\texttt{merge\_g400\_t06} & 2.0 & 79.14 & 72.00 & 83.59 & 434 \\
\texttt{merge\_g400\_t07} & 2.0 & 79.19 & 72.09 & 83.64 & 436 \\
\texttt{merge\_g450\_mtoff\_t08} & off & 84.19 & 81.51 & 91.96 & 256 \\
\texttt{merge\_g450\_mtoff\_t085} & off & 84.10 & 81.33 & 91.66 & 272 \\
\texttt{merge\_g600\_mtoff\_t085} & off & 84.11 & 81.34 & 91.67 & 273 \\
\bottomrule
\end{tabular}
}
\end{table}

Table~\ref{tab:supp_merge_sport} gives the three merge configurations
discussed in Sec.~\ref{sec:exp_merge} with their per-sport breakdown.

\begin{table}[t]
\centering
\caption{\textbf{Offline tracklet merge, per-sport breakdown} on
  SportsMOT val (pooled TrackEval columns as in Sec.~\ref{sec:exp_merge};
  per-sport HOTA is the unweighted sequence mean). Bold: selected
  configuration (best worst-sport HOTA).}
\label{tab:supp_merge_sport}
\resizebox{\columnwidth}{!}{
\begin{tabular}{lccccccc}
\toprule
 & \multicolumn{4}{c}{Overall (pooled)} & \multicolumn{3}{c}{HOTA by sport} \\
\cmidrule(lr){2-5}\cmidrule(lr){6-8}
Configuration & HOTA$\uparrow$ & AssA$\uparrow$ & IDF1$\uparrow$ & IDSW$\downarrow$ & Bask. & Foot. & Voll. \\
\midrule
No merge (online tracker) & 77.97 & 69.78 & 81.48 & 514 & 79.28 & 75.41 & 81.59 \\
Merge, motion gate on ($g{=}100$, $\tau{=}0.70$) & 79.20 & 72.11 & 83.65 & 436 & 80.55 & 76.67 & 82.38 \\
\textbf{Merge, gate off ($g{=}300$, $\tau{=}0.85$)} & \textbf{84.12} & \textbf{81.37} & \textbf{91.72} & \textbf{273} & \textbf{85.57} & \textbf{83.03} & \textbf{83.86} \\
\bottomrule
\end{tabular}
}
\end{table}

\section{Transfer and runtime tables}
\label{sec:supp_transfer}

Table~\ref{tab:transfer} gives the merge-transfer comparison
referenced in Sec.~\ref{sec:exp_merge} (pooled TrackEval scores on both
benchmarks, different embedding models), and Table~\ref{tab:fps} the
runtime breakdown referenced in Sec.~\ref{sec:exp_setup}.

\begin{table}[t]
\centering
\caption{\textbf{Merge transfer across benchmarks.} The same
  offline merge ($\tau{=}0.85$, $g{=}300$, motion gate off)
  applied to each benchmark's online val output, with different
  embedding models (sports OSNet vs.\ FastReID SBS-S50):
  appearance-only re-identification with the motion veto
  disabled is benchmark-general. All cells are pooled TrackEval
  scores, matching the merge table in the main paper.}
\label{tab:transfer}
\resizebox{\columnwidth}{!}{
\begin{tabular}{lcccccc}
\toprule
 & \multicolumn{3}{c}{HOTA$\uparrow$} & \multicolumn{3}{c}{AssA$\uparrow$} \\
\cmidrule(lr){2-4}\cmidrule(lr){5-7}
Benchmark (val) & online & $+$merge & $\Delta$ & online & $+$merge & $\Delta$ \\
\midrule
SportsMOT & 77.97 & 84.12 & $+$6.15 & 69.78 & 81.37 & $+$11.59 \\
DanceTrack & 56.86 & 58.73 & $+$1.87 & 41.65 & 44.70 & $+$3.04 \\
\bottomrule
\end{tabular}
}
\end{table}

\begin{table}[t]
\centering
\caption{\textbf{Runtime} on a single RTX 3060 Laptop GPU
  (SportsMOT val), all mechanisms active. Deferred commitment costs
  ${\approx}0.6$\,ms/frame, visible only in the network-free
  microbenchmark (base configuration 234.1\,FPS); with any per-frame
  network the cost disappears into noise. Every stage exceeds the
  25\,FPS broadcast rate.}
\label{tab:fps}
\resizebox{\columnwidth}{!}{
\begin{tabular}{lc}
\toprule
Pipeline stage & FPS$\uparrow$ \\
\midrule
Association only (cached detections + CMC, no ReID) & 204.8 \\
Tracker + live OSNet re-identification (ReID), excl.\ image decode & 64.8 \\
Full pipeline (incl.\ JPEG decode) & 50.7 \\
\bottomrule
\end{tabular}
}
\end{table}

\section{Historical baselines omitted from the SportsMOT test comparison}
\label{sec:supp_baselines}

For completeness, three older SportsMOT test rows omitted from
Table~\ref{tab:test_comparison} for space, as reported in the SportsMOT
paper~\cite{cui2023sportsmot} (Table~3, train$+$val detector, linear
interpolation with maximum gap 20): ByteTrack~\cite{zhang2022bytetrack}
HOTA 64.1 / DetA 78.5 / AssA 52.3 / MOTA 95.9 / IDF1 71.4 / IDs 3089;
OC-SORT~\cite{cao2023ocsort} 73.7 / 88.5 / 61.5 / 96.5 / 74.0 / 2728;
MixSort-OC~\cite{cui2023sportsmot} 74.1 / 88.5 / 62.0 / 96.5 / 74.4 /
2781. None is cited by number in the main text.

\section{Shared-detection study}
\label{sec:supp_shared_det}

Table~\ref{tab:supp_shared_det} lists the seven-tracker comparison
summarized in Sec.~\ref{sec:exp_sota}: every tracker consumes the
same cached YOLOX-X detections on SportsMOT val, outputs are scored
raw (no interpolation), so the differences are attributable to
association alone. \method{} is best in HOTA, AssA, IDF1, and
identity switches while having the lowest DetA of the seven.

\begin{table}[t]
\centering
\caption{\textbf{Shared-detection study} on SportsMOT val: seven
  SORT-family trackers run on identical cached YOLOX-X detections
  (raw tracker outputs, no interpolation; pooled TrackEval). With the
  detector held fixed, the spread is association alone.}
\label{tab:supp_shared_det}
\resizebox{\columnwidth}{!}{
\begin{tabular}{lcccccc}
\toprule
Tracker & HOTA$\uparrow$ & DetA$\uparrow$ & AssA$\uparrow$ & MOTA$\uparrow$ & IDF1$\uparrow$ & IDs$\downarrow$ \\
\midrule
\textbf{\method{} (ours)} & \textbf{77.71} & \textbf{86.94} & \textbf{69.50} & \textbf{95.39} & \textbf{81.35} & \textbf{552} \\
BoT-SORT~\cite{aharon2022botsort} & 76.49 & 88.54 & 66.12 & 95.68 & 77.43 & 731 \\
ByteTrack~\cite{zhang2022bytetrack} & 75.01 & 87.73 & 64.17 & 95.04 & 75.76 & 812 \\
Deep OC-SORT~\cite{maggiolino2023deep_ocsort} & 74.99 & 88.05 & 63.91 & 95.07 & 75.99 & 787 \\
StrongSORT~\cite{du2023strongsort} & 73.43 & 87.28 & 61.82 & 94.17 & 73.80 & 1854 \\
OC-SORT~\cite{cao2023ocsort} & 72.51 & 86.77 & 60.63 & 93.64 & 73.26 & 1208 \\
SORT~\cite{bewley2016sort} & 71.75 & 87.30 & 59.02 & 94.14 & 70.87 & 1191 \\
\bottomrule
\end{tabular}
}
\end{table}

\section{Per-sequence results}
\label{sec:supp_per_seq}

Table~\ref{tab:supp_per_seq} reports per-sequence HOTA/AssA/IDF1 of
the selected configuration on all 45 SportsMOT val sequences.

\begin{table*}[t]
\centering
\caption{\textbf{Per-sequence results} of the selected
  configuration (online $+$ interpolation $+$ merge,
  $\tau{=}0.85$, $g{=}300$, gate off) on SportsMOT val.}
\label{tab:supp_per_seq}
\resizebox{0.98\textwidth}{!}{
\begin{tabular}{llccc@{\hspace{2em}}llccc}
\toprule
Sequence & Sport & HOTA & AssA & IDF1 & Sequence & Sport & HOTA & AssA & IDF1 \\
\midrule
\texttt{v\_00HRwkvvjtQ\_c001} & basketball & 86.95 & 84.66 & 94.79 & \texttt{v\_9MHDmAMxO5I\_c006} & volleyball & 83.72 & 82.31 & 93.36 \\
\texttt{v\_00HRwkvvjtQ\_c003} & basketball & 91.60 & 91.72 & 99.03 & \texttt{v\_9MHDmAMxO5I\_c009} & volleyball & 81.92 & 77.01 & 89.73 \\
\texttt{v\_00HRwkvvjtQ\_c005} & basketball & 89.01 & 89.12 & 98.88 & \texttt{v\_BgwzTUxJaeU\_c008} & basketball & 83.78 & 78.84 & 93.04 \\
\texttt{v\_00HRwkvvjtQ\_c007} & basketball & 83.48 & 79.89 & 94.80 & \texttt{v\_BgwzTUxJaeU\_c012} & basketball & 81.58 & 76.65 & 91.76 \\
\texttt{v\_00HRwkvvjtQ\_c008} & basketball & 89.81 & 90.10 & 99.32 & \texttt{v\_BgwzTUxJaeU\_c014} & basketball & 84.48 & 79.28 & 92.70 \\
\texttt{v\_00HRwkvvjtQ\_c011} & basketball & 86.75 & 84.68 & 94.76 & \texttt{v\_G-vNjfx1GGc\_c004} & football & 83.35 & 79.93 & 89.52 \\
\texttt{v\_0kUtTtmLaJA\_c004} & volleyball & 80.29 & 73.33 & 87.78 & \texttt{v\_G-vNjfx1GGc\_c008} & football & 83.67 & 83.83 & 90.55 \\
\texttt{v\_0kUtTtmLaJA\_c005} & volleyball & 83.31 & 80.08 & 91.30 & \texttt{v\_G-vNjfx1GGc\_c600} & football & 82.11 & 84.62 & 94.76 \\
\texttt{v\_0kUtTtmLaJA\_c006} & volleyball & 78.64 & 75.05 & 85.25 & \texttt{v\_G-vNjfx1GGc\_c601} & football & 81.35 & 80.70 & 90.67 \\
\texttt{v\_0kUtTtmLaJA\_c007} & volleyball & 88.03 & 88.83 & 98.38 & \texttt{v\_ITo3sCnpw\_k\_c007} & football & 78.83 & 75.82 & 85.68 \\
\texttt{v\_0kUtTtmLaJA\_c008} & volleyball & 88.77 & 88.69 & 97.82 & \texttt{v\_ITo3sCnpw\_k\_c010} & football & 85.19 & 82.09 & 92.66 \\
\texttt{v\_0kUtTtmLaJA\_c010} & volleyball & 83.87 & 82.66 & 92.95 & \texttt{v\_ITo3sCnpw\_k\_c011} & football & 82.77 & 81.38 & 89.64 \\
\texttt{v\_2QhNRucNC7E\_c017} & football & 88.44 & 87.50 & 94.66 & \texttt{v\_ITo3sCnpw\_k\_c012} & football & 79.38 & 72.82 & 84.02 \\
\texttt{v\_4-EmEtrturE\_c009} & volleyball & 93.95 & 94.11 & 99.84 & \texttt{v\_cC2mHWqMcjk\_c007} & volleyball & 88.04 & 86.57 & 95.75 \\
\texttt{v\_4r8QL\_wglzQ\_c001} & basketball & 90.66 & 89.91 & 98.40 & \texttt{v\_cC2mHWqMcjk\_c008} & volleyball & 86.99 & 85.27 & 94.21 \\
\texttt{v\_5ekaksddqrc\_c001} & basketball & 82.17 & 77.10 & 92.27 & \texttt{v\_cC2mHWqMcjk\_c009} & volleyball & 79.35 & 69.44 & 83.18 \\
\texttt{v\_5ekaksddqrc\_c002} & basketball & 87.41 & 84.40 & 96.34 & \texttt{v\_dw7LOz17Omg\_c053} & football & 89.47 & 88.97 & 98.26 \\
\texttt{v\_5ekaksddqrc\_c003} & basketball & 80.54 & 72.93 & 88.37 & \texttt{v\_dw7LOz17Omg\_c067} & football & 91.49 & 92.40 & 98.77 \\
\texttt{v\_5ekaksddqrc\_c004} & basketball & 83.73 & 77.74 & 91.44 & \texttt{v\_i2\_L4qquVg0\_c006} & football & 78.37 & 76.68 & 84.94 \\
\texttt{v\_5ekaksddqrc\_c005} & basketball & 81.56 & 74.43 & 87.26 & \texttt{v\_i2\_L4qquVg0\_c007} & football & 83.37 & 83.11 & 90.83 \\
\texttt{v\_9MHDmAMxO5I\_c002} & volleyball & 78.91 & 73.22 & 86.64 & \texttt{v\_i2\_L4qquVg0\_c009} & football & 81.22 & 77.05 & 88.46 \\
\texttt{v\_9MHDmAMxO5I\_c003} & volleyball & 84.74 & 80.88 & 91.65 & \texttt{v\_i2\_L4qquVg0\_c010} & football & 76.36 & 71.45 & 81.99 \\
\texttt{v\_9MHDmAMxO5I\_c004} & volleyball & 77.33 & 80.17 & 83.65 &  &  &  &  &  \\
\bottomrule
\end{tabular}
}
\end{table*}

\end{document}